%% file: main.tex
\documentclass[11pt]{article}

\usepackage[preprint]{acl}

\usepackage{times}
\usepackage{latexsym}

\usepackage[T1]{fontenc}

\usepackage[utf8]{inputenc}

\usepackage{microtype}

\usepackage{inconsolata}

\usepackage{graphicx}

\usepackage{amsmath}
\usepackage{times}
\usepackage{framed}
\usepackage{ragged2e}
\usepackage{bm}
\usepackage{soul}
\usepackage{latexsym}
\usepackage{subfigure}
\usepackage{amsthm}
\usepackage{graphicx}
\usepackage{float}
\usepackage{longtable}
\usepackage{booktabs} 
\usepackage{multirow}
\usepackage{multicol}
\usepackage{amssymb}
\usepackage{dashbox}%
\usepackage{multirow}
\usepackage{textcomp}
\usepackage{tcolorbox}
\usepackage{bbding}
\usepackage{bbm}
\usepackage[ruled, vlined, linesnumbered]{algorithm2e}
\usepackage{mathtools}
\usepackage{adjustbox}
\usepackage[ruled,vlined]{algorithm2e}
\usepackage{color, soul}
\usepackage{pifont}
\usepackage{array}
\newcolumntype{P}[1]{>{\centering\arraybackslash}p{#1}}
\newcolumntype{M}[1]{>{\centering\arraybackslash}m{#1}}
\usepackage{cleveref}
\crefname{section}{§}{§§}
\Crefname{section}{§}{§§}
\crefname{figure}{Figure}{Figure}
\Crefname{figure}{Figure}{Figure}
\crefname{table}{Table}{Table}
\Crefname{table}{Table}{Table}
\usepackage{tabularx}
\usepackage{booktabs}
\usepackage[table,xcdraw]{xcolor}
\usepackage{makecell}

\newcommand\ourmas{\textsc{TrajDebug}\xspace}
\newcommand{\tautwobench}{$\tau^2$-Bench}
\newcommand\ourdata{\textsc{TrajErrBench}\xspace}

\usepackage{tcolorbox}
\tcbuselibrary{listings, breakable}

\title{\ourmas: Tracing Error Lifecycle to Identify Critical Failures \\in Long-Horizon Agent Trajectories}

\author{
\normalfont Yunjia Qi\textsuperscript{1},
Zehua Yin\textsuperscript{1},
Xintong Shi\textsuperscript{1},
Hao Peng\textsuperscript{1}, \\
Songyuanyi Lu\textsuperscript{3},
Yixian Liu\textsuperscript{3},
Richeng Xuan\textsuperscript{3},
Yuhong Liu\textsuperscript{3},
Zhichao Hu\textsuperscript{3,*}, \\
Xiaozhi Wang\textsuperscript{2},
Lei Hou\textsuperscript{1},
Bin Xu\textsuperscript{1,*},
Juanzi Li\textsuperscript{1} \\
\textsuperscript{1}Department of Computer Science and Technology, BNRist,
Tsinghua University \\
\textsuperscript{2}Shenzhen International Graduate School,
Tsinghua University \\
\textsuperscript{3}Tencent Hunyuan \\
\textsuperscript{*}Corresponding authors \\
\texttt{qyj23@mails.tsinghua.edu.cn}
}

\begin{document}
\maketitle
\begin{abstract}

LLM-based agentic systems have shown remarkable capabilities in complex domains, while suffering from cascading errors and difficulty in debugging. 
Critical error detection aims to locate the earliest error step in a failed trajectory that is responsible for the final failure.
However, progress faces two main challenges. First, long trajectories make it difficult to identify individual errors, since the evidence for judging a step may be scattered across distant instructions, observations, and prior context. Second, failed trajectories often contain multiple local errors whose downstream effects differ: some are repaired, some remain harmless, and only some contribute to the final failure. In this work, we propose \ourmas, an error-lifecycle tracing framework that addresses long-trajectory error discovery with multi-granularity history compression and evidence-based error identification, and supports critical attribution by tracing each error’s resolution status and terminal impact.
We further construct \ourdata, a benchmark of $486$ manually annotated failed trajectories from \tautwobench and SWE-Bench Pro, covering realistic tool-use and coding scenarios. Experiments across diverse agent benchmarks show that \ourmas achieves the best overall performance over existing baselines, and application studies further demonstrate that its diagnoses provide actionable feedback for improving downstream agent success. 
We will release the code and data to facilitate further research\footnote{\url{https://github.com/THU-KEG/TrajDebug}}.
\end{abstract}

\input{sections/01.introduction}

\input{sections/02.preliminary}
\input{sections/03.method}

\input{sections/04.experiments}

\input{sections/05.relatedwork}
\input{sections/06.conclusion}

\section*{Acknowledgments}
This work is supported by the 2025 Tencent Rhino-Bird Joint Research Program (JR2025TEG013).

\bibliography{custom}
\appendix

\clearpage

\input{sections/appendix}

\end{document}

%% file: sections/01.introduction.tex
\section{Introduction}

Large language model (LLM)-based agentic systems have shown remarkable capabilities in complex, real-world domains, such as software engineering~\citep{jimenez2024swebench,deng2025swepro}, scientific discovery~\citep{mialon2024gaia}, and multi-turn customer service workflows~\citep{barres2025tau2}.
These trajectories often span hundreds of reasoning, action, and observation steps, and a terminal failure may trace back to an early mistake that propagates through subsequent decisions, making the trajectory difficult to debug and improve.
\emph{Critical error detection}~\citep{zhang2025whoandwhen} addresses this 
difficulty by locating the earliest error step in a failed trajectory that is causally 
linked to the task failure, thereby supporting trajectory repair and system 
reliability analysis.

\begin{figure}[t]
\centering
\includegraphics[width=0.85\linewidth]{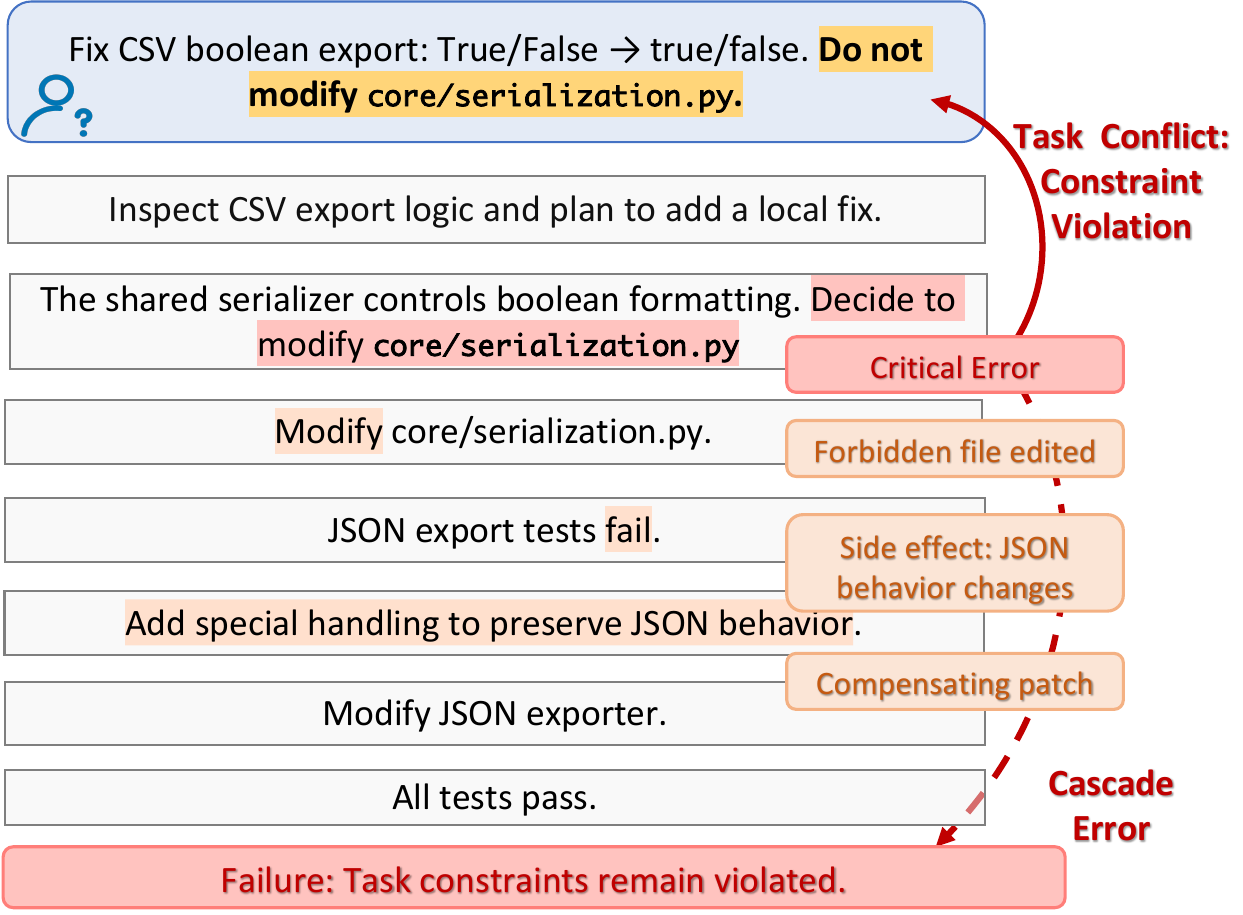}
\caption{Critical error detection requires grounding errors in long-range context and distinguishing the failure-responsible error from multiple coexisting errors.}
\label{fig:fig1}
\vspace{-0.3cm}
\end{figure}

Critical error detection faces two key challenges in long, complex agent trajectories.
First, \textit{long trajectories make it difficult to identify individual errors.}
Complex tasks often require extended execution processes that interleave planning, reasoning, tool use, environment feedback, and repair attempts over many steps~\citep{yao2023tree,yao2022react}.
As a result, judging whether a step is erroneous may require relating it to evidence scattered across distant task instructions, observations, and prior trajectory context~\citep{peng2023does}.
For example, in Figure~\ref{fig:fig1}, the erroneous decision conflicts with a task constraint stated at the beginning of the trajectory, requiring long-range task context to identify the error.
Second, \textit{failed trajectories often contain multiple local errors whose downstream effects differ, obscuring which error is critical to the final failure.}
A failed trajectory may contain many local errors~\citep{cemri2026multi}, some are resolved, some are inconsequential, and others are merely downstream symptoms of earlier mistakes~\citep{fan2026agentprocessbench}. 
Thus, the critical error is not necessarily the first local error observed in the trajectory, nor the temporally closest error to the terminal failure.

Existing methods often address only part of this problem.
Current taxonomy-based and constraint-based methods focus on identifying candidate erroneous steps in long trajectories~\citep{zhu2025agentdebug,barke2026agentrx}, while causal-diagnostic methods use causal structures or counterfactual attribution to filter propagated errors~\citep{wang2026flat,li2026codetracer}.
However, candidate-error methods often leave criticality to a holistic final judgment, which is brittle when many local errors coexist.
Causal attribution methods model dependencies among errors, but their judgments can remain ambiguous when later feedback and repair attempts alter the effects of earlier errors.

To address these challenges, we propose \ourmas, an evidence-grounded error-lifecycle tracing framework for critical error detection.
\ourmas decomposes the task into three stages: error trigger detection, error state classification, and critical attribution.
First, to identify individual errors in long trajectories, \ourmas uses multi-granularity history compression and detects error triggers as wrong commitments grounded in conflicts with task instructions, trajectory history, environment feedback, or the agent's own reasoning.
This turns error discovery from holistic diagnosis into an auditable evidence-verification problem, reducing hallucinated or weakly grounded diagnoses~\citep{gao2023enabling}.
Second, to address the ambiguity caused by multiple local errors with different downstream effects, \ourmas groups related triggers into error instances by their shared violated reference, such as a task constraint or prior observation.
It then classifies the state of each instance by tracking whether the wrong commitment is resolved, and whether it leaves an observable terminal footprint, such as an irreversible state change, a persistent task violation, or a substantial recovery cost.
These state classifications distinguish errors that are repaired or remain harmless from those that stay relevant to the final failure.
Finally, \ourmas selects the critical error step from evidence-backed instances with terminal relevance, rather than judging over the full trajectory.

To evaluate critical error detection under realistic scenarios and complement existing benchmarks~\citep{zhang2025whoandwhen,zhu2025agentdebug}, we construct \ourdata, 
a benchmark of $486$ manually annotated failed trajectories: $400$ from \tautwobench~\citep{barres2025tau2}, covering diverse tool-use and user-interaction scenarios, and $86$ from SWE-Bench Pro~\citep{deng2025swepro}, covering long-horizon coding trajectories with an average length of about $119.7$ steps.

Experiments on existing agent benchmarks and \ourdata show that \ourmas achieves the best overall performance over advanced LLM prompting baselines and existing diagnostic systems.
Length-based analysis shows that \ourmas remains more robust on long-horizon trajectories.
To assess whether critical-error diagnoses can improve future agent behavior, we explore two inference-time application scenarios.
In the first, each failed trajectory is diagnosed and converted into targeted guidance before re-executing the same task, improving success by $10.80\%$ on average.
In the second, diagnoses from a small set of historical failures are aggregated into reusable failure memory and transferred to held-out tasks, yielding a $5.70\%$ average improvement.
These results suggest that \ourmas can serve not only as a diagnostic tool, but also as a practical interface for converting failed executions into actionable experience for improving long-horizon agents.

%% file: sections/02.preliminary.tex
\section{Pilot Study}
\label{sec:pilot}

This section first formalizes critical error step detection (\S~\ref{sec:pilot:prelim}) and then provides empirical motivation for the two challenges discussed above. We examine how trajectory length is associated with critical error localization difficulty (\S~\ref{sec:rq1}) and how local errors evolve within failed trajectories (\S~\ref{sec:rq2}). See Appendix~\ref{sec:appendix:pilot} for experiment and annotation details.

\subsection{Task Formulation}
\label{sec:pilot:prelim}
Let a trajectory be $\tau = (x_1, x_2, \ldots, x_N)$, where each step $x_t$ contains the agent's reasoning, action, or environment response. A trajectory is \textit{failed} if its final state does not satisfy the task goal.
Following~\citet{zhang2025whoandwhen}, a step $x_t$ contains a decisive error if replacing its action with a correct one, while leaving all prior steps unchanged and allowing the remainder of the trajectory to unfold correctly, would have turned the failure into a success.
The \textbf{critical error step} is the earliest decisive error step, identifying the origin of the failure rather than its downstream propagation.

\subsection{Error Detection under Growing Context}
\label{sec:rq1}

\begin{figure}[t]
    \centering
    \includegraphics[width=0.9\linewidth]{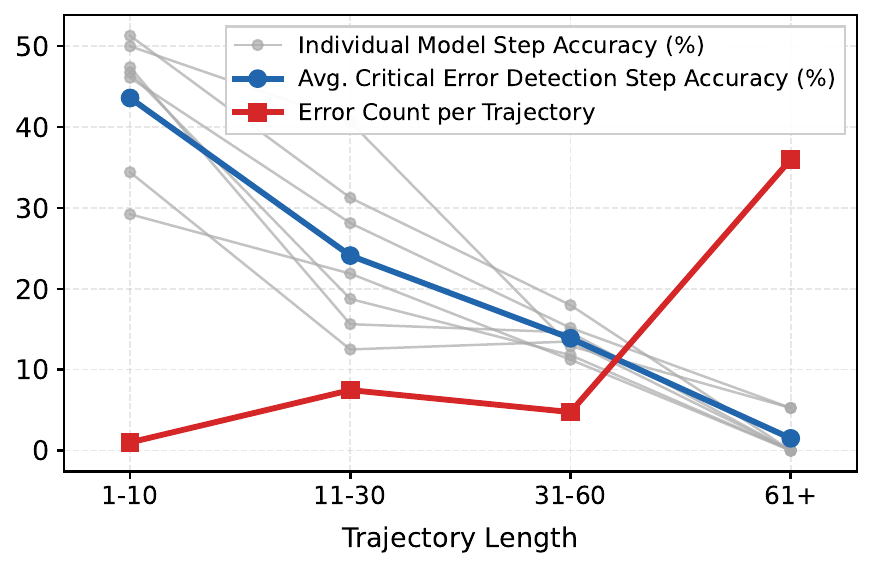}
    \caption{Critical error detection accuracy and local error density across trajectory length buckets.}
    \label{fig:critical_position}
    \vspace{-0.3cm}
\end{figure}

We group trajectories from WhoAndWhen~\citep{zhang2025whoandwhen} and AgentDebugBench~\citep{zhu2025agentdebug} by trajectory length, and directly prompt LLMs to identify the annotated critical error step in each failed trajectory, and use critical-step detection accuracy as an indicator of the difficulty introduced by growing context.
We evaluate seven widely used advanced models and report the average accuracy across models to reveal the overall trend.
Figure~\ref{fig:critical_position} shows that detection accuracy decreases as trajectory length increases, suggesting that long-horizon context makes critical error detection harder.
This trend is consistent with prior observations that longer trajectories are associated with lower task success rates~\citep{qi2026agentif}.

\subsection{Trajectory Dynamics of Local Errors}
\label{sec:rq2}

We further sample $50$ failed trajectories from these benchmarks and annotate local error steps with three annotators, with details in Appendix~\ref{sec:appendix:pilot}.
The sampled trajectories contain $381$ local errors in total, averaging $7.62$ per failed trajectory, while each trajectory has only one critical error.
Figure~\ref{fig:critical_position} shows that the average number of local errors increases with trajectory length, indicating that longer trajectories expose a denser set of plausible but non-critical candidates.
To understand how non-critical errors evolve, we exclude the critical errors and analyze the remaining $331$ local errors.
We find that $205$ ($61.9\%$) are later repaired by the agent, while $104$ ($31.4\%$) persist until the final outcome.
Only $22$ ($6.6\%$) are unrepaired yet dormant: for example, a step may miscount four search results as two, but no later decision depends on the count.
These dynamics show that critical error detection must distinguish local error occurrence from its downstream state and terminal impact, motivating \ourmas in \S~\ref{sec:method:lifecycle}.

%% file: sections/03.method.tex
\section{\ourmas}

\begin{figure*}[t]
    \centering
    \includegraphics[width=0.9\linewidth]{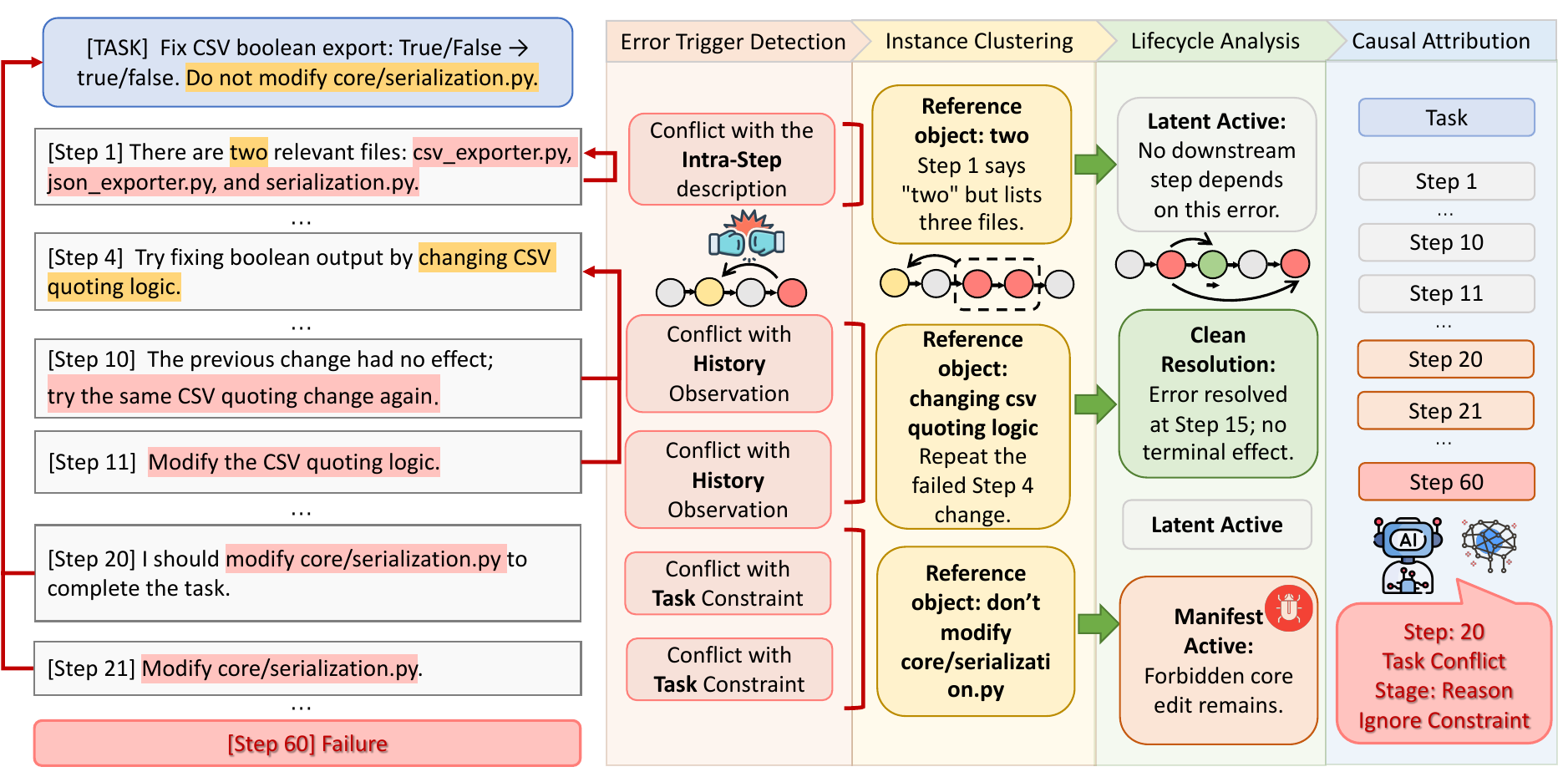} 
    \caption{An overview of the framework of \ourmas.
    }
    \vspace{-0.3cm}
    \label{fig:pipeline}
\end{figure*}

Figure~\ref{fig:pipeline} shows the overall framework of \ourmas.
Given a failed trajectory $\tau$, \ourmas first constructs multi-granularity trajectory views to preserve local evidence while compressing distant context (\S\ref{sec:method:compression}).
\ourmas then follows an error-lifecycle tracking perspective and decomposes critical error detection into three stages.
First, \ourmas detects evidence-grounded error \textit{triggers} at individual steps (\S\ref{sec:method:detect}).
Second, it groups related triggers into object-anchored error \textit{instances} and classifies each instance's error state, based on whether the error is resolved or remains active and whether it leaves an observable terminal footprint (\S\ref{sec:method:lifecycle}).
Finally, it attributes the final failure to the critical candidate through candidate-set-guided causal attribution (\S\ref{sec:method:select}).
This turns critical error detection from an end-to-end judgment over the full trajectory into three more controlled steps: finding evidence-backed local errors, classifying their error states, and selecting the causally responsible step from the remaining candidates.
More details and examples are in Appendix~\ref{sec:appendix:case} and Appendix~\ref{sec:appendix:prompts}.

\subsection{Multi-Granularity Compression}
\label{sec:method:compression}
To reduce the long-context burden while preserving verifiable evidence, we use LLM to construct three views for each step:
(1) \textit{high-detail} view keeps the original instruction, action, observation, and locally relevant reasoning snippets needed for evidence verification;
(2) \textit{medium-detail} view summarizes the step's main intent, action, and state update;
(3) \textit{low-detail} view records only coarse progress, salient entities, and unresolved commitments.
Each stage then retrieves the finest view required for its decision, using high-detail views for local verification and compressed views for distant context.

\subsection{Error Trigger Detection}
\label{sec:method:detect}

For each step, given the trajectory prefix up to that step, the first stage extracts per-step atomic error triggers.
A \textbf{trigger} $e=(t,c,p,q_w,q_r)$ denotes a local evidence-grounded mismatch at step $t$, where $q_w$ is an erroneous commitment expressed in the current step, $q_r$ is the violated reference, $c$ is the reference category, and $p$ is the execution phase.
The violated reference $q_r$ may come from the task instruction, prior trajectory, environment feedback, or the same step.
To prevent subjective drift and hallucinated diagnoses, each trigger must satisfy a \textbf{verbatim evidence} condition: both $q_w$ and $q_r$ must be explicitly citable; otherwise, the trigger is discarded.
A step may contain multiple triggers when it violates different references or expresses different erroneous commitments.
For each inspected step, the detector uses the high-detail current step and task instruction, medium-detail previous two steps, and low-detail views for the remaining history.
This preserves local evidence for verification while keeping long-range context compact.

We organize triggers along two axes.
The first axis is the \textbf{reference category $c$}, which specifies the source of the violated reference $q_r$:
(1) \textit{Task Conflict}, where $q_r$ comes from the task instruction, such as a constraint;
(2) \textit{History Conflict}, where $q_r$ comes from prior trajectory context, such as a tool output, or established trajectory fact;
and (3) \textit{Intra-Step Conflict}, where $q_r$ comes from the current step itself, such as an earlier claim.
We additionally use (4) \textit{Environment Anomaly} as an exogenous category when the agent action is reasonable but the environment response is abnormal.
The second axis is the \textbf{execution phase $p$}, which records where the mismatch surfaces in the agent process~\citep{deshpande2025trail,zhu2025agentdebug}: \textit{planning}, \textit{reasoning}, \textit{action}, \textit{observation}, or \textit{verification} (detailed in Appendix~\ref{sec:appendix:prompts}).
The reference category $c$, together with the violated reference $q_r$, determines the reference object $O$ used for instance clustering, while the execution phase $p$ characterizes the agent-side mechanism for fine-grained analysis.

\subsection{Error State Classification}
\label{sec:method:lifecycle}

The second stage groups per-step triggers into error \textbf{instances} and classifies the state of each instance.
Since the same wrong commitment may appear across multiple steps~\cite{xie2026spark}, such as a task-constraint violation surfacing in planning, action, and verification, we define an instance as $E=(\mathcal{E},O)$, where $\mathcal{E}$ contains triggers that violate the same reference object $O$.
This avoids overcounting repeated manifestations of the same error.
Each instance thus represents one continuous episode of holding a wrong commitment to a concrete reference object.

For each instance, the state classifier makes two evidence-backed judgments: whether the wrong commitment is \textbf{resolved} or remains \textbf{active}, and whether it leaves an observable \textbf{terminal footprint}.
An instance is resolved only if later steps explicitly revisit $O$ and provide citable evidence that supersedes the wrong commitment; otherwise, it remains active.
A terminal footprint is assigned when there is evidence of: 
(i) an \textit{irreversible} state change, such as submitting a wrong order; 
(ii) a \textit{semantic} footprint, where the wrong commitment is reflected in later steps or the terminal state, such as a persistent violation of a task constraint; 
or (iii) \textit{budget debt}, where the error is resolved only after consuming more than $k\%$ of the trajectory. 
In implementation, we set $k=50$, as exceeding half the trajectory mostly leaves an insufficient budget for recovery.

Combining these two judgments yields four error states: \textbf{Clean Resolution} (resolved, no footprint), \textbf{Costly Resolution} (resolved, budget-debt footprint), \textbf{Manifest Active} (active, irreversible or semantic footprint), and \textbf{Latent Active} (active, no footprint).
We retain terminal-relevant instances as candidates:
\begin{equation}
\begin{aligned}
\mathcal{F}(\tau)=\{E:\operatorname{state}(E) \in \\
\{\operatorname{Costly Resolution},\operatorname{Manifest Active}\}\}.
\end{aligned}
\end{equation}
Each candidate is passed to the final stage with its origin step, state label, footprint channel, and supporting evidence; final criticality is decided by the attribution stage.

\subsection{Candidate-Set-Guided Causal Attribution}
\label{sec:method:select}

This stage executes candidate-set-guided causal attribution.
Given the candidate set $\mathcal{F}(\tau)$, this attribution head selects the candidate instance whose first step best explains the terminal failure.
A simple earliest-candidate rule is insufficient because temporal order alone does not determine failure responsibility.
For example, an earlier budget-debt instance should not be selected over a later semantic instance unless the wasted budget plausibly prevented the agent from avoiding the later failure.
Therefore, we use an LLM as the final attribution head.
Each candidate is provided with its first step, state label, and supporting verbatim evidence, and the LLM selects the critical error step.

\begin{figure}[t]
    \centering
    \includegraphics[width=\linewidth]{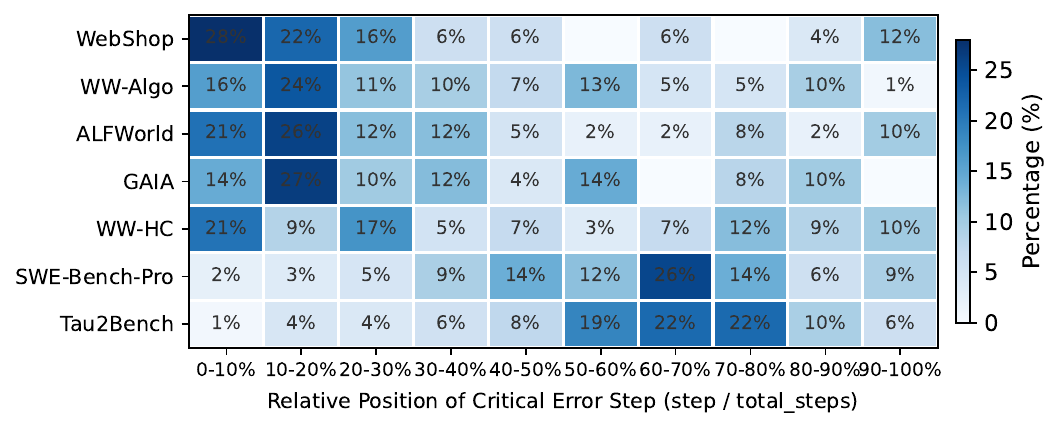}
    \caption{Relative position of the critical error step within each trajectory in \ourdata and others.}
    \vspace{-0.3cm}
\label{fig:benchmark:critical_position}
\end{figure}

\section{\ourdata}
\label{sec:data}

Existing critical-error benchmarks remain limited in scale, domain diversity, or trajectory complexity~\cite{zhang2025whoandwhen,zhu2025agentdebug,barke2026agentrx,li2026codetracer}.
We construct \ourdata, a benchmark of $486$ manually annotated failed trajectories from two complementary domains: $400$ \tautwobench\ trajectories~\citep{barres2025tau2} and $86$ SWE-Bench Pro trajectories~\citep{deng2025swepro}, averaging $29.3$ and $119.7$ steps, respectively.
Following WhoAndWhen~\citep{zhang2025whoandwhen}, annotators identify the earliest evidence-grounded mistake responsible for the final failure, while separating locally wrong but later repaired mistakes from failure-responsible ones.
Each trajectory is annotated by three annotators, and labels with at least two-way agreement are used as ground truth; this majority-vote protocol yields usable labels for $98.2\%$ of \tautwobench\ trajectories and $91.9\%$ of SWE-Bench Pro trajectories.
The critical-error step labels reach almost-perfect agreement on \tautwobench\ (Fleiss' $\kappa{=}0.91$) and substantial agreement on the longer, code-heavy SWE-Bench Pro trajectories (Fleiss' $\kappa{=}0.67$); details are in Appendix~\ref{sec:appendix:ourdata}.

Figure~\ref{fig:benchmark:critical_position} shows that critical errors in \ourdata often occur in the mid-to-late.
This reflects the shared structure of \ourdata: agents first gather task-relevant information before making decisive decisions, through user interaction in \tautwobench\ and repository exploration in SWE-Bench Pro.
Such extended information gathering shifts many critical errors later and makes localization harder, as relevant evidence is accumulated across earlier context.
We also annotate each critical error by contradiction reference category and execution phase, with full statistics in Appendix~\ref{sec:appendix:ourdata}.
In \tautwobench, critical errors are almost evenly split between task conflicts ($52.4\%$) and history conflicts ($46.3\%$), while SWE-Bench Pro is dominated by task conflicts ($73.4\%$) with fewer history conflicts ($24.1\%$). Reasoning is the dominant failure phase in both domains ($60.7\%$ and $57.1\%$), suggesting that failures mainly arise from misinterpreting accumulated context in agentic trajectories.

%% file: sections/04.experiments.tex
\section{Experiments}
\label{sec:experiments}

\begin{table*}[t]
  \centering
  \adjustbox{max width=\linewidth}{%
  \begin{tabular}{lcccccccc}
  \toprule
  \multirow{2}{*}{\textbf{Method}} &
  \multicolumn{3}{c}{\textbf{AgentDebugBench}} &
  \multicolumn{2}{c}{\textbf{WhoAndWhen}} &
  \multicolumn{2}{c}{\textbf{\ourdata}} &
  \multirow{2}{*}{\textbf{AVG}} \\
  \cmidrule(lr){2-4}\cmidrule(lr){5-6}\cmidrule(lr){7-8}
  & ALFWorld & GAIA & WebShop & Hand-Crafted & Algorithm & \tautwobench & SWE-Bench Pro &  \\
  \midrule
  \textit{Avg. Trajectory Length} & $60.00$ & $28.68$ & $51.84$ & $51.60$ & $8.72$ & $29.27$ & $119.70$ & $49.97$ \\
  \midrule
  \rowcolor{gray!15}
  \multicolumn{9}{l}{\textit{Direct Prompting}} \\
  GLM-5.1            & $15.00$ & $26.00$ & $26.00$ & $15.52$ & $42.86$ & $46.75$ & $6.98$ & $25.59$ \\
  Gemini-3.1-Pro     & $19.00$ & $30.61$ & $\mathbf{32.00}$ & $20.69$ & $46.82$ & $52.50$ & $17.44$ & $31.29$ \\
  DeepSeek-V4-Pro    & $15.00$ & $34.69$ & $24.00$ & $18.97$ & $42.06$ & $48.25$ & $12.79$ & $27.97$ \\
  Claude Sonnet 4.6  & $13.00$ & $30.61$ & $18.00$ & $12.07$ & $29.37$ & $42.75$ & $15.12$ & $22.99$ \\
  GPT-5.4            & $11.00$ & $20.41$ & $14.00$ & $13.79$ & $28.57$ & $41.50$ & $10.46$ & $19.96$ \\
  Claude Opus 4.6    & $10.00$ & $34.69$ & $24.00$ & $\mathbf{27.59}$ & $46.83$ & $45.25$ & $17.44$ & $29.40$ \\
  Qwen3-235B-A22B    & $9.00$  & $33.33$ & $18.00$ & $17.24$ & $42.06$ & $42.75$ & $17.44$ & $25.69$ \\
  \rowcolor{gray!15}
  \multicolumn{9}{l}{\textit{Multi-Agent System}} \\
  AgentDebugger & $21.00$ & $36.00$ & $20.00$ & $12.07$ & $40.48$ & $26.00$ & $10.47$ & $23.72$ \\
  CHIEF              & $6.00$  & $26.00$ & $10.00$ & $18.97$ & $41.27$ & $26.82$ & $2.32$  & $18.77$ \\
  AgentRX    & $8.00$  & $26.53$ & $18.00$     & $25.86$     & $41.27$     & $36.25$     & $5.81$     & $23.10$ \\
  \midrule
  \rowcolor{gray!15}
  \textbf{\ourmas\ (Ours)} & $\mathbf{26.00}$ & $\mathbf{38.78}$ & $26.00$ & $22.41$ & $\mathbf{48.41}$ & $\mathbf{52.75}$ & $\mathbf{24.41}$ & $\mathbf{34.11}$ \\
  \bottomrule
  \end{tabular}%
  }
\caption{Main results on critical error detection across multiple benchmarks. All methods operate without task ground truth and receive only the failed trajectory. \textbf{Bold} values indicate the best result per column.}
\label{tab:main_results}
\end{table*}

\subsection{Experimental Settings}

\paragraph{Datasets and Evaluation Metric.}
We evaluate our approach across benchmarks in various domains.
\textbf{WhoAndWhen}~\citep{zhang2025whoandwhen} contains a hand-crafted subset of $58$ trajectories and an algorithm-generated subset of $126$ trajectories drawn from GAIA and AssistantBench.
\textbf{AgentDebugBench}~\citep{zhu2025agentdebug} provides $100$ ALFWorld and $50$ each from GAIA and WebShop trajectories.
One GAIA trajectory has a null ground-truth critical-error label and is therefore excluded from evaluation, leaving $869$ evaluated trajectories in total.
\textbf{\ourdata} (\S~\ref{sec:data}) adds $400$ \tautwobench\ and $86$ SWE-Bench Pro trajectories, extending evaluation to substantially longer tool-use and software-engineering rollouts.
Following prior benchmark~\citep{zhang2025whoandwhen,zhu2025agentdebug}, we report \textbf{exact critical-step accuracy}, where a prediction is correct only if it matches the annotated step, and evaluate all methods in a realistic setting where they use only the failed trajectory and failure signal, without success traces, gold outcomes, or auxiliary debugging signals.

\paragraph{Baselines.}
We compare with two families of baselines.
Motivated by the effectiveness of direct prompting reported in prior work~\citep{zhang2025whoandwhen,banerjee2025did}, our first family is \textbf{Direct Prompting}, which queries an LLM for the critical error step from the full trajectory under a unified instruction; we instantiate it with seven widely-used advanced models: GLM-5.1~\citep{zeng2026glm5}, Gemini-3.1-Pro~\citep{team2023gemini}, DeepSeek-V4-Pro~\citep{guo2025deepseek}, Claude Sonnet 4.6~\citep{claude46sonnet}, Claude Opus 4.6~\citep{claude46opus}, GPT-5.4~\citep{gpt4}, and Qwen3-235B-A22B-Thinking~\citep{yang2025qwen3}. 
The second family is \textbf{Multi-Agent Systems}, which decomposes the task across multiple LLM calls:
\textit{AgentDebugger}~\citep{zhu2025agentdebug} prompts the judge with a phase-level error taxonomy and produces a structured diagnosis;
\textit{CHIEF}~\citep{wang2026flat} parses the trajectory into a hierarchical causal graph and selects the critical step via counterfactual backtracking against a virtual oracle;
\textit{AgentRX}~\citep{barke2026agentrx} first flags steps that violate explicit task or schema constraints and then asks an LLM to select the critical error.

\paragraph{Implementation.}
We implement \ourmas\ and all multi-agent baselines with Qwen3-235B-A22B-Thinking~\citep{yang2025qwen3} for a fair comparison, while direct-prompting baselines use the LLMs listed in Table~\ref{tab:main_results}.
All methods use temperature $0$ for reproducibility, and multi-agent baselines follow official prompts and pipelines where available.
Implementation details are in Appendix~\ref{sec:appendix:experiment}.

\subsection{Main Results}

All experimental results are presented in Table~\ref{tab:main_results}.
We have the following observations:
(1) \ourmas\ achieves the best overall average in our benchmark suite.
It obtains the highest macro-average accuracy ($34.11\%$), outperforming both direct prompting with frontier models and all multi-agent baselines on average.
The absolute accuracies remain low across all methods, highlighting the difficulty of identifying the exact critical step from only the failed trajectory.
Under this challenging setting, \ourmas\ improves over direct prompting with the same backbone from $25.69\%$ to $34.11\%$ ($+8.42$), demonstrating the effectiveness of our framework.
(2) \ourmas\ generalizes well across heterogeneous agent domains.
It ranks first on five of seven datasets, spanning embodied decision-making, open-domain information seeking, user-facing tool use, and long-horizon software engineering.
In contrast, multi-agent baselines show uneven cross-domain performance(e.g., CHIEF drops to $2.32\%$ on SWE-Bench Pro and AgentRX to $8.00\%$ on ALFWorld), suggesting that their pipelines transfer less consistently across agent scenarios.
These results demonstrate that \ourmas's evidence-grounded trigger detection and error state classification offer robust advantages that generalize across diverse agent scenarios.
(3) \ourmas\ remains effective on the longest-horizon benchmarks.
On ALFWorld ($60.00$ steps on average) and SWE-Bench Pro ($119.70$ steps), where evidence is distributed across many steps and multiple errors often coexist, \ourmas\ attains the best accuracy, with gains of $+13.00\%$ and $+6.97\%$ over the corresponding direct baselines.
These results suggest that evidence-grounded trigger detection and error state classification help narrow critical-step localization in long trajectories with complex failure dynamics.

\subsection{Ablation Study}
Table~\ref{tab:ablation} ablates the three key components of \ourmas\ on \tautwobench\ and SWE-Bench Pro. Replacing multi-granularity compression with the same hard-truncation strategy used by direct prompting causes the largest drop, reducing AVG by $21.02$ points. This shows that preserving evidence from different history granularities is essential for long trajectories, rather than merely increasing the number of LLM calls.
Removing error trigger detection or error state classification lowers AVG by $4.99$ and $7.31$ points, respectively, showing that these two downstream components are also useful and non-redundant.
On \tautwobench, both ablations reach the same accuracy, suggesting that the trigger taxonomy and state classification stage act as orthogonal filters of comparable strength on tool-use trajectories. On SWE-Bench Pro, removing state classification reduces accuracy to $16.28$, below the direct-prompting baseline of $17.44$, whereas removing only the taxonomy still achieves $20.93$. This pattern supports our motivation in \S~\ref{sec:pilot}: long-horizon trajectories contain many detected errors that are cleanly resolved or dormant, making final attribution more difficult.
State-based filtering narrows attribution to errors with terminal footprint.

\begin{table}[t]
  \centering
  \adjustbox{max width=\linewidth}{%
  \begin{tabular}{lccc}
  \toprule
  \textbf{Method} & \tautwobench & SWE-Bench Pro & AVG \\
  \midrule
  \textbf{\ourmas\ (Ours)}        & $\mathbf{52.75}$ & $\mathbf{24.41}$ & $\mathbf{38.58}$ \\
  w/o Multi-Granularity Compression & $20.00$ & $15.12$ & $17.56$ \\
  w/o Evidence-Grounded Triggers              & $46.25$ & $20.93$ & $33.59$ \\
  w/o Error State Classification   & $46.25$ & $16.28$ & $31.27$ \\
  w/o All Components      & $42.75$ & $17.44$ & $30.10$ \\
  \bottomrule
  \end{tabular}%
  }
  \caption{Ablation study of \ourmas.}
  \label{tab:ablation}
\end{table}

\subsection{Length Analysis}

\begin{figure}[t]
    \centering
    \includegraphics[width=0.95\linewidth]{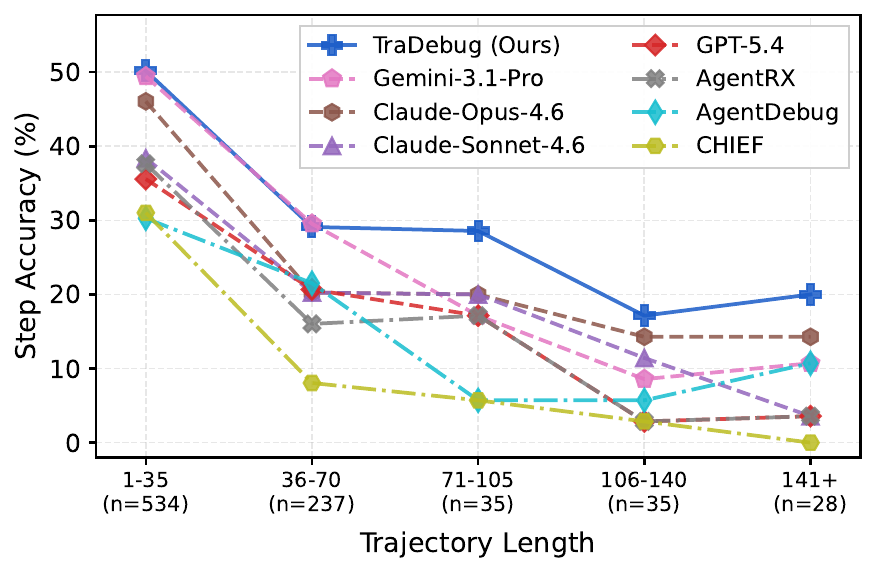}
    \caption{Accuracy across trajectory length buckets.}
    \label{fig:length}
    \vspace{-0.3cm}
\end{figure}

Figure~\ref{fig:length} shows critical-step detection accuracy across trajectory length buckets. All baselines degrade sharply as trajectories grow, dropping from $35$--$50\%$ on short trajectories to below $15\%$ on long ones. In contrast, \ourmas remains substantially more stable, retaining over $20\%$ accuracy on the longest bucket where the best baseline drops to around $14\%$. 
This suggests that \ourmas is still affected by increasing trajectory length, but evidence-grounded error detection and state-based candidate filtering help it degrade less sharply and yield larger gains on long-horizon trajectories.

\section{Application: Critical Error Detection as Feedback for Agent Improvement}
\label{sec:applications}

Beyond diagnostic accuracy, a critical error detector is useful only if its outputs translate into measurable agent improvement.
We evaluate \ourmas\ as a feedback source for the agent in two deployment scenarios on \tautwobench~\citep{barres2025tau2} and a sampled set of $100$ SWE-Bench Verified~\citep{jimenez2024swebench}, covering both an idealized per-trajectory repair setting and a more realistic cross-trajectory transfer setting.
In both scenarios, the actor is GLM-5.1, all detectors use Qwen3-235B-A22B-Thinking as the backbone, and detector-produced feedback is injected into the actor's system prompt.
We compare \ourmas\ against two baselines: \textit{Vanilla}, which directly prompts with the failed trajectory to find the critical error step and write feedback, and \textit{Self-Reflection}~\citep{renze2024self}, which generates a reflection over the trajectory before producing the feedback.
Details are in Appendix~\ref{sec:appendix:application}.

\begin{table}[t]
\centering
\small
\adjustbox{max width=\linewidth}{%
\begin{tabular}{lccc}
\toprule
\textbf{Method} & \textbf{Airline} & \textbf{Retail} & \textbf{SWE-Bench} \\
\midrule
\multicolumn{4}{l}{\emph{Per-Trajectory Repair (Oracle Failure)}} \\
\midrule
Initial          & $78.00$ & $84.21$ & $72.00$ \\
Self-Reflection  & $84.00$ \footnotesize{(+$6.00$)}  & $93.86$ \footnotesize{(+$9.65$)} & $78.00$ \footnotesize{(+$6.00$)}  \\
Vanilla Debug         & $88.00$ \footnotesize{(+$10.00$)} & $93.86$ \footnotesize{(+$9.65$)}  & $80.00$ \footnotesize{(+$8.00$)}  \\
\textbf{\ourmas} & $\mathbf{90.00}$ \footnotesize{(\textbf{+$12.00$})} & $\mathbf{95.61}$ \footnotesize{(\textbf{+$11.40$})} & $\mathbf{81.00}$ \footnotesize{(\textbf{+$9.00$})} \\
\midrule
\multicolumn{4}{l}{\emph{Failure-Memory Transfer (Realistic)}} \\
\midrule
Initial          & $78.78$ & $84.21$ & $71.64$ \\
Self-Reflection  & $81.81$ \footnotesize{(+$3.03$)}  & $82.89$ \footnotesize{(-$1.32$)}  & $74.63$ \footnotesize{(+$2.99$)} \\
Vanilla Debug         & $75.75$ \footnotesize{(-$3.03$)}  & $88.15$ \footnotesize{(+$3.94$)} & $71.64$ \footnotesize{(+$0.00$)} \\
\textbf{\ourmas} & $\mathbf{84.84}$ \footnotesize{(\textbf{+$6.06$})} & $\mathbf{90.78}$ \footnotesize{(\textbf{+$6.57$})} & $\mathbf{76.12}$ \footnotesize{(\textbf{+$4.48$})} \\
\bottomrule
\end{tabular}%
}
\caption{Task success rate (\%) when detector outputs are used as feedback for GLM-5.1.}
\label{tab:application}
\end{table}

\paragraph{Per-Trajectory Repair under Oracle Failure.}
We assume oracle access to the success label of each rollout: for every failed trajectory, the detector produces a feedback, which is injected into the system prompt before re-executing the same task with the same actor.
As shown in the upper block of Table~\ref{tab:application}, \ourmas\ delivers the largest gains across all three settings, lifting the average success rate from $78.07$ to $88.87$, surpassing Vanilla Debug and Self-Reflection.
The gap shows that more accurate critical-error localization produces more actionable feedback, beyond what holistic prompting or generic reflection can offer.

\paragraph{Failure-Memory Transfer to Held-Out Trajectories.}
We further test a realistic deployment in which the detector is applied to only a small slice of historical failures and the resulting feedback must generalize to unseen tasks.
For each subset, we split the trajectories per-dataset into a $1/3$ \emph{memory-construction split} and a $2/3$ \emph{held-out evaluation split}.
On the memory-construction split, the detector produces feedback on each failed trajectory; all feedbacks are aggregated into a memory pool, which is then injected as a whole into the actor's system prompt when evaluating on the held-out split.
As shown in the lower block of Table~\ref{tab:application}, despite no per-instance oracle, \ourmas\ still yields the largest improvement, whereas Vanilla and Self-Reflection show less stable transfer and can even reduce performance in some cases.
This indicates that the feedback induced by \ourmas\ encodes transferable failure patterns rather than instance-specific fixes, suggesting that critical error detection can serve as a low-cost source of reusable experience for long-horizon agents.

%% file: sections/05.relatedwork.tex
\section{Related Work}

\paragraph{Trajectory analysis and process supervision.}
Recent work analyzes agent executions beyond final answers, including process reward models that assign step-level scores~\citep{xi2026agentprm,fan2026agentprocessbench} and studies that summarize recurring failure modes such as coordination, tool-use, and planning errors~\citep{cemri2026multi,ma2024agentboard,lu2024agentlens,mulian2026agentfixer,ma2025diagnosing}.
However, step-level scores do not isolate the failure-responsible step, and failure-mode taxonomies identify error categories rather than the decisive step in a specific trajectory.

\paragraph{Critical error attribution.}
A recent line studies fine-grained failure attribution across multi-agent reasoning~\citep{zhang2025whoandwhen,chen2026elephant,in2026mpbench,banerjee2025did}, embodied and web environments~\citep{zhu2025agentdebug}, and software engineering~\citep{deshpande2025trail,li2026codetracer}.
Methodologically, prompt-based approaches inspect raw or compressed trajectories with an LLM judge~\citep{zhang2025whoandwhen,banerjee2025did}, but their diagnoses can be weakly grounded in evidence;
taxonomy- and constraint-based approaches narrow the search space with predefined error categories or invariants~\citep{zhu2025agentdebug,barke2026agentrx}, yet miss errors that fall outside the schema;
replay- and spectrum-based methods localize errors by re-executing or perturbing the trajectory~\citep{ge2025famas}, but require deterministic environments;
graph-based methods model dependencies among trajectory elements~\citep{wang2026flat,zhang2025graphtracer,zhang2025agentracer,li2026codetracer,wang2026agenttrace}, but use connectedness as a proxy for terminal responsibility, leaving each error's resolution status and terminal impact implicit.
In contrast, \ourmas grounds error triggers in verbatim evidence, classifies error states, and performs attribution over terminal-relevant candidates, addressing both long-context error identification and ambiguity among multiple local errors.

\paragraph{Benchmarks for failure attribution.}
Existing benchmarks are typically tied to a single domain and modest trajectory length: WhoAndWhen~\citep{zhang2025whoandwhen}, TraceElephant~\citep{chen2026elephant}, and MPBench~\citep{in2026mpbench} target multi-agent dialogues; AgentDebug~\citep{zhu2025agentdebug} covers embodied and web tasks; and TRAIL~\citep{deshpande2025trail} and CodeTracer~\citep{li2026codetracer} focus on code agents.
\ourdata complements them with $486$ manually annotated long-horizon trajectories from customer-service tool use~\citep{barres2025tau2} and repository-scale software engineering~\citep{deng2025swepro}, with up to ${\sim}120$ steps per trajectory, exposing critical errors under long-range evidence dependencies and multiple coexisting local errors.

%% file: sections/06.conclusion.tex
\section{Conclusion}

We presented \ourmas, a three-stage error-lifecycle tracing framework for critical error detection.
It combines error trigger detection, error state classification, and candidate-set-guided attribution to identify errors in long trajectories and distinguish failure-responsible errors among multiple errors.
We also introduced \ourdata, a benchmark of $486$ manually annotated failed trajectories from realistic tool-use and coding scenarios.
Experiments show that \ourmas outperforms LLM prompting and diagnostic baselines, and application studies demonstrate its value for trajectory repair and transferable failure memory.
These results position critical error detection as a promising interface for inference-time agent improvement.

\section{Limitations}
We discuss the limitations of our work here, including two main aspects:
(1) Although \ourmas constrains judgments with verbatim evidence and structured candidates, it still relies on LLMs for error interpretation and final attribution. Its predictions may therefore be affected by the model's reasoning ability, domain knowledge, and calibration. To reduce the concern that the gains merely come from using a stronger judge, we implement \ourmas with an open-source backbone rather than the strongest proprietary models; even under this setting, \ourmas outperforms direct prompting with frontier models and all multi-agent baselines on average.
(2) As a staged pipeline, \ourmas may inherit false negatives from earlier trigger detection or state classification stages, causing the final candidate set to miss the true critical error. To mitigate this, the attribution stage allows an out-of-set prediction only under strict evidence requirements: the model must explain why the retained candidates do not account for the failure and provide the missing trigger, violated reference, origin step, and terminal footprint.

\section{Ethical Considerations}
We discuss three ethical considerations. (1) \textit{Intellectual property.} We respect the licenses of all artifacts used in this work, including datasets, models, and code repositories. We will release \ourmas, the associated code, and \ourdata under the MIT license\footnote{https://opensource.org/license/mit}. (2) \textit{Intended use and risk control.} \ourmas is designed to trace the error lifecycle for critical error detection in failed agent trajectories. The data used in this work is anonymized to the best of our knowledge. Since the framework may still produce incorrect predictions due to model and method limitations, users should manually verify important diagnoses before using them in high-stakes settings. (3) \textit{AI assistance.} We used AI to refine the wording of some sentences.

%% file: sections/appendix.tex
\section{Pilot Study Details}
\label{sec:appendix:pilot}

\paragraph{Data sources and annotators.}
For Pilot Study~\ref{sec:pilot}, we randomly sample $50$ failed trajectories from WhoAndWhen~\citep{zhang2025whoandwhen} and AgentDebugBench~\citep{zhu2025agentdebug}, drawing $10$ trajectories from each subset of the two benchmarks, with $1{,}373$ steps in total.
The critical-error label of each trajectory is inherited from the source dataset, since both benchmarks already provide ground-truth critical-error annotations for their failed trajectories; we therefore only annotate per-step local errors and their downstream states.
All annotations are produced by three computer-science undergraduates who completed a calibration training round on a separate set of trajectories before working on the pilot data.

\paragraph{Detector evaluation (\S\ref{sec:rq1}).}
For the detector comparison in \S\ref{sec:rq1}, we evaluate seven advanced models: GLM-5.1~\citep{zeng2026glm5}, Gemini-3.1-Pro~\citep{team2023gemini}, DeepSeek-V4-Pro~\citep{guo2025deepseek}, Claude Sonnet 4.6~\citep{claude46sonnet}, Claude Opus 4.6~\citep{claude46opus}, GPT-5.4~\citep{gpt4}, and Qwen3-235B-A22B~\citep{yang2025qwen3}.
All models are queried with the same direct-prompt template used by the direct prompting baselines (Table~\ref{tab:appendix:direct_prompting_baseline}) at temperature $0$, and a prediction is counted as correct only when the predicted step index exactly matches the critical error step provided by the source dataset.

\paragraph{Local error step annotation (\S\ref{sec:rq1}, \S\ref{sec:rq2}).}
We recruit a pool of $20$ annotators in total, all of whom completed a calibration training round on a held-out set of trajectories.
Annotators were compensated at rates consistent with the prevailing market standards for comparable annotation work.
Three annotators independently inspect each step.
Following the annotation principles established in prior process-supervision work~\citep{fan2026agentprocessbench}, each annotator is required to (i) judge each step strictly along the agent's own reasoning, action, and observation, without using their own task knowledge to fill in unstated steps; (ii) mark a step as erroneous only when the error can be tied to an explicit, objective trigger, such as a violated task constraint, a contradicted observation, or an internally inconsistent claim; and (iii) write a short, reproducible reason that cites the specific evidence supporting the judgment. Vague reasons (e.g., ``the agent seems off-track'') are rejected and the step is not added to the error set.
A step enters the error set only when all three annotators independently mark it as erroneous and each provides an evidence-grounded reason; disagreements are kept as non-errors to avoid inflating the error set with subjective calls.
This protocol yields $381$ local error steps.
The Fleiss' $\kappa$ on error identification is $0.538$, indicating moderate agreement and confirming that, even under an evidence-grounded protocol, identifying step-level errors in long agent trajectories remains genuinely hard for humans.

\paragraph{Downstream-state annotation of non-critical local errors (\S\ref{sec:rq2}).}
For the downstream-state analysis in \S\ref{sec:rq2}, the same three annotators independently classify each of the $331$ non-critical local errors into one of three outcomes: \textit{repaired}, \textit{persistent}, or \textit{unrepaired yet dormant}.
We attach an evidence requirement to each label so that state judgments remain reproducible.
A \textit{repaired} label follows the resolution criterion in \S\ref{sec:method:lifecycle}: a later step must explicitly revisit the same reference object and supersede the wrong commitment, for example by satisfying the violated constraint, acting on the corrected state, or retracting the prior wrong claim; the annotator must cite the specific repairing step.
A \textit{persistent} label requires the annotator to write an explicit reason together with the evidence from a later step that still uses or relies on the wrong commitment introduced at the error step.
An \textit{unrepaired yet dormant} label is reserved for cases where no later step depends on the erroneous content, for example a step that miscounts four search results as two when no subsequent reasoning or action references the count; the annotator must state why the wrong content is never used downstream.
The boundary between \textit{persistent} and \textit{unrepaired yet dormant} is therefore decided by the presence of a downstream reference: persistent requires a citable later usage of the wrong content, while dormant requires explaining its absence.
A label is kept only when all three annotators independently agree, following the same unanimous-agreement criterion used for local error identification.
The Fleiss' $\kappa$ on downstream-state labels is $0.378$, lower than the agreement on error identification and consistent with the view that state judgments require additional reasoning over downstream references and are therefore harder to align across annotators.

\section{\ourdata Details}
\label{sec:appendix:ourdata}

\subsection{Dataset Construction}
\label{sec:appendix:dataset}

We construct \ourdata from two complementary domains: $400$ rollouts on the airline and retail tasks of \tautwobench~\citep{barres2025tau2}, produced by DeepSeek-Reasoner, GPT-5, Qwen3-Max
and $86$ rollouts on SWE-Bench Pro~\citep{deng2025swepro}, produced by Claude-Opus-4-6~\citep{claude46opus}, Kimi-K2.6-Coding~\citep{team2026kimi}, Gemini-3.1-Pro~\citep{team2023gemini}, Grok-4.20-Beta~\citep{xai2025-grok4-fast-model-card}, and GPT-5.4~\citep{gpt4} from the OpenHands scaffold~\citep{wang2025openhands}.
The two subsets average $29.3$ and $119.7$ steps per trajectory, respectively.

\subsection{Annotation Details}
\label{sec:appendix:annotation}

Our annotation protocol follows the decisive-error annotation practice of WhoAndWhen~\citep{zhang2025whoandwhen}. 
Annotators are given the task instruction, the full failed trajectory, and the final failure outcome, and are asked to identify the earliest step that directly causes the failure under the counterfactual definition in \S~\ref{sec:pilot:prelim}.
They do not need to execute an actual intervention; instead, they make an evidence-grounded attribution judgment from the trajectory, as in prior failure-attribution annotation.

Since both benchmarks contain long and difficult failure trajectories, we first generate model-assisted pre-annotations to help annotators understand the likely failure points.
Specifically, we ask GPT-5.4, Gemini-3.1-Pro, and Claude-Opus-4-6 to annotate each trajectory.
Each model is given the full trajectory, the ground-truth task specification, and the failure information returned by the original benchmark evaluation.
For \tautwobench, this includes the unmet database constraints reported by the task environment; for SWE-Bench Pro, this includes the details of the failing test cases.
Annotators then receive the same full information, together with the three model pre-annotations, as reference material during annotation.
The pre-annotations are used only to support trajectory understanding; the final labels are determined by human annotators following the guideline below.

\paragraph{Agreement between pre-annotations and final labels.}
To quantify potential anchoring from model-assisted pre-annotation, Table~\ref{tab:appendix:preannotation_agreement} reports the exact-match rate between each model's proposed critical step and the final human-majority label.
Agreement varies substantially across models and subsets, and no single model consistently determines the final labels.
Moreover, the pre-annotation setting provides benchmark failure details to aid human annotation, whereas all evaluated direct-prompting baselines receive only the failed trajectory and failure signal, without gold outcomes or auxiliary debugging information.

\begin{table}[h]
\centering
\small
\caption{Exact-match agreement between model-assisted pre-annotations and final human-majority labels.}
\label{tab:appendix:preannotation_agreement}
\adjustbox{max width=\linewidth}{%
\begin{tabular}{llc}
\toprule
Subset & Pre-annotation model & Exact match \\
\midrule
\multirow{3}{*}{\tautwobench\ ($N{=}400$)}
& Claude Opus 4.6 & $64.5\%$ \\
& Gemini 3.1 Pro  & $76.8\%$ \\
& GPT-5.4         & $53.2\%$ \\
\midrule
\multirow{3}{*}{SWE-Bench Pro ($N{=}86$)}
& Claude Opus 4.6 & $48.8\%$ \\
& Gemini 3.1 Pro  & $31.4\%$ \\
& GPT-5.4         & $55.8\%$ \\
\bottomrule
\end{tabular}
}
\end{table}

The standardized guideline contains three principles.
First, an annotated step must contain an explicit mistake supported by the available evidence, such as violating the task instruction, contradicting a previous observation, misusing a tool result, or making an internally inconsistent claim.
Second, the mistake must be failure-responsible: annotators should not choose local mistakes that are later corrected or errors whose downstream effect never affects the final outcome.
Third, if multiple failure-responsible mistakes are present, annotators choose the earliest such step, following the principal-cause convention used by WhoAndWhen~\citep{zhang2025whoandwhen}.
Each annotator also writes a short explanation for the selected step so that disagreements can be checked against concrete evidence.

\paragraph{Annotator pool and labelling protocol.}
We recruit a pool of $20$ annotators in total, all of whom completed a calibration training round on a held-out set of trajectories covering both \tautwobench\ and SWE-Bench Pro before working on the released data.
Annotators were compensated at rates consistent with the prevailing market standards for comparable annotation work.
Because decisive-error localization is inherently ambiguous in long agent trajectories, each trajectory is independently annotated by $3$ annotators and the final label is taken by majority vote: a label is kept when at least $2$ of the $3$ annotators agree.
We apply the same three-annotator majority-vote protocol to the reference-category labels (\textit{Task Conflict}, \textit{History Conflict}, and \textit{Intra-Step Conflict}) and to the execution-phase labels (\textit{planning}, \textit{reasoning}, \textit{action}, \textit{observation}, \textit{verification}).

\paragraph{Inter-annotator agreement.}
Table~\ref{tab:appendix:annotation_agreement} reports inter-annotator agreement for the critical-error step, the execution-phase label, and the reference-category label on each subset of \ourdata.
On \tautwobench, the critical-error step reaches almost-perfect agreement (Fleiss' $\kappa{=}0.91$), while the execution-phase and reference-category labels reach substantial and fair agreement respectively.
For SWE-Bench Pro, we initially annotate $110$ trajectories, including ambiguous cases on which no two annotators select the same critical step.
The released benchmark and all experiments use only the $86$ trajectories for which at least two of three annotators agree on the critical step; Table~\ref{tab:appendix:annotation_agreement} reports agreement on this retained set.
Agreement is lower than on \tautwobench, reflecting the longer and more code-heavy trajectories: the critical-error step reaches substantial agreement (Fleiss' $\kappa{=}0.674$), while execution-phase and reference-category agreement is moderate and fair, respectively.
Despite this, $98.2\%$ of \tautwobench\ trajectories and $91.9\%$ of SWE-Bench Pro trajectories receive a reference-category label on which at least $2$ of $3$ annotators agree, indicating that the majority-vote protocol yields a usable label on the overwhelming majority of trajectories.

\begin{table}[h]
\centering
\small
\caption{Inter-annotator agreement on \ourdata. ``Pairwise'' denotes pairwise agreement.}
\label{tab:appendix:annotation_agreement}
\adjustbox{max width=\linewidth}{%
\begin{tabular}{llcc}
\toprule
Subset & Label & Fleiss' $\kappa$ & Pairwise \\
\midrule
\multirow{3}{*}{\tautwobench} 
& Critical-error step  & $0.91$ & $91.2\%$ \\
& Execution phase      & $0.73$ & $84.0\%$ \\
& Reference category   & $0.41$ & $68.9\%$ \\
\midrule
\multirow{3}{*}{SWE-Bench Pro}
& Critical-error step  & $0.67$ & $66.7\%$ \\
& Execution phase      & $0.42$ & $61.8\%$ \\
& Reference category   & $0.38$ & $52.4\%$ \\
\bottomrule
\end{tabular}
}
\end{table}

\subsection{Distribution of Critical Errors}
\label{sec:appendix:critical_distribution}

\begin{figure*}[t]
    \centering
    \includegraphics[width=0.95\linewidth]{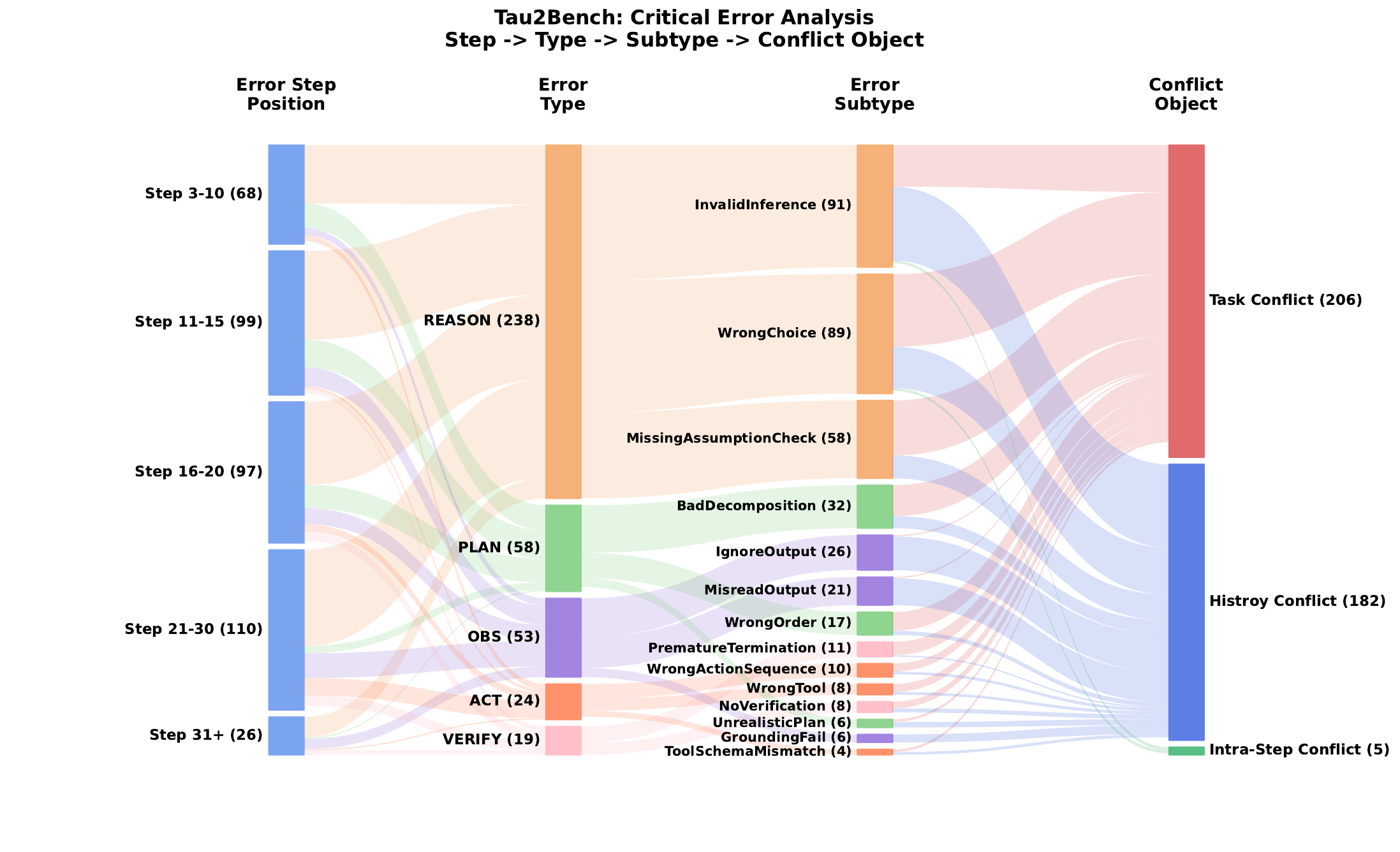}
    \caption{Distribution of critical errors in \tautwobench. Flows go from execution phase (left) to error subtype (middle) to reference category (right).}
    \label{fig:appendix:sankey_tau2}
\end{figure*}

\begin{figure*}[t]
    \centering
    \includegraphics[width=0.95\linewidth]{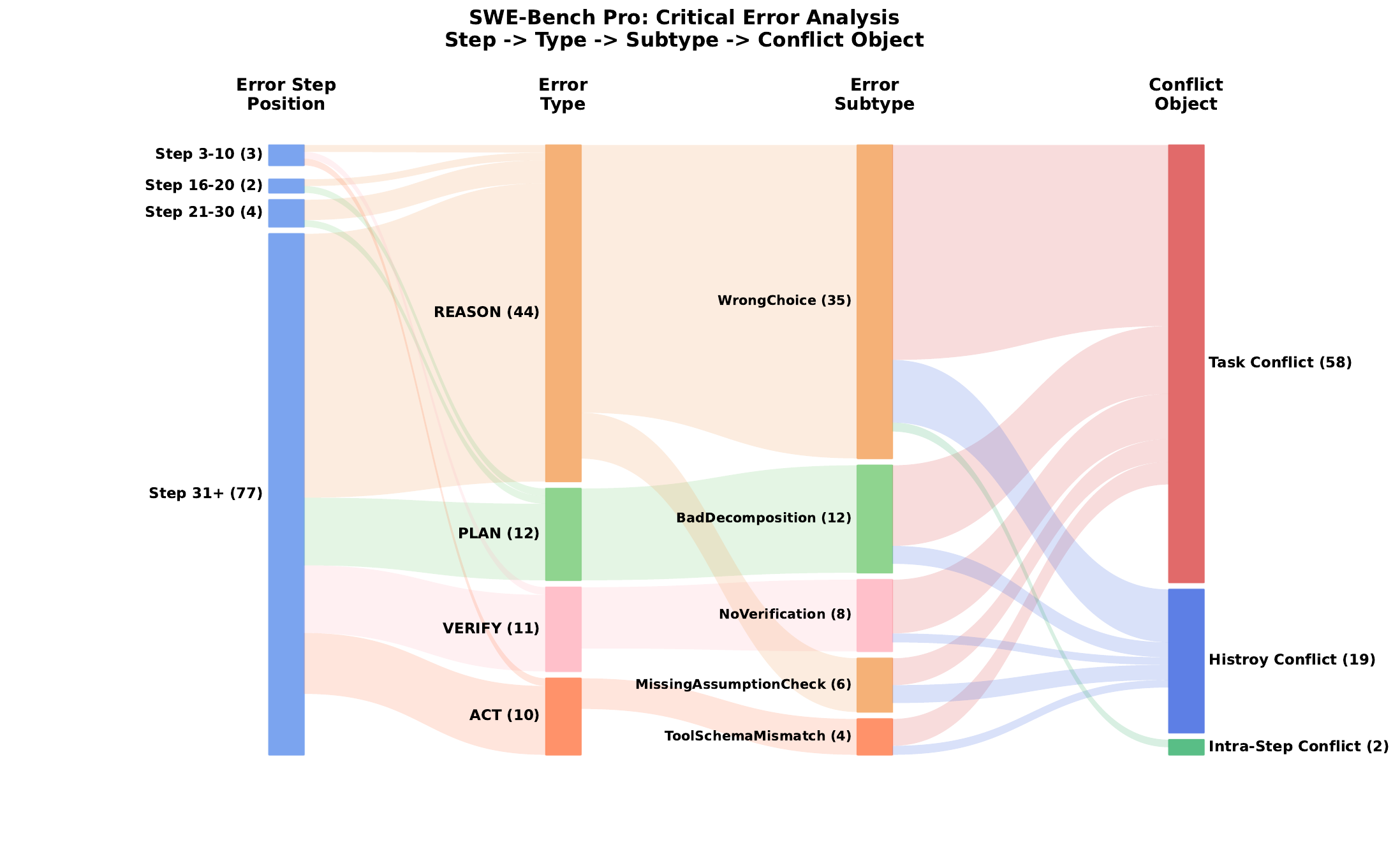}
    \caption{Distribution of critical errors in SWE-Bench Pro. Flows go from execution phase (left) to error subtype (middle) to reference category (right).}
    \label{fig:appendix:sankey_swebench}
\end{figure*}

Figures~\ref{fig:appendix:sankey_tau2} and~\ref{fig:appendix:sankey_swebench} show the joint distribution of critical errors over the execution phase, error subtype, and reference category for the \tautwobench\ and SWE-Bench Pro subsets of \ourdata respectively.
The flows in each diagram are read from left (execution phase) to right (reference category), with the middle column showing the fine-grained error subtype.

\paragraph{Execution phase.}
Among critical errors with an agreed execution-phase label, \tautwobench\ is heavily skewed toward reasoning ($60.7\%$), followed by planning ($14.8\%$) and observation ($13.5\%$), with action ($6.1\%$) and verification ($4.8\%$) contributing the rest.
This indicates that on tool-use tasks, critical errors more often arise when the agent interprets accumulated context than when it issues an isolated action.
In SWE-Bench Pro, reasoning is still the largest phase ($57.1\%$), followed by planning ($15.6\%$), verification ($14.3\%$), and action ($13.0\%$).
This shift is consistent with the patch-and-test nature of software-engineering trajectories: the agent spends comparatively more time editing code and validating it, and a larger share of failures appears in those phases.

\paragraph{Error subtype.}
The middle column reveals which fine-grained agent behaviour is most failure-prone within each phase.
On \tautwobench, the top three subtypes \texttt{reason.InvalidInference} ($22.8\%$), \texttt{reason.WrongChoice} ($22.2\%$), and \texttt{reason.MissingAssumptionCheck} ($14.5\%$) jointly account for nearly $60\%$ of critical errors, indicating that the dominant failure mode is inference over already-observed context rather than acting on incorrect observations.
On SWE-Bench Pro, the distribution becomes more peaked: \texttt{reason.WrongChoice} alone accounts for $40.7\%$ of critical errors, followed by \texttt{plan.BadDecomposition} ($14.0\%$) and \texttt{verify.NoVerification} ($9.3\%$), suggesting that failures cluster around selecting the wrong implementation path and skipping verification before submission.

\paragraph{Reference category.}
On the right column, both subsets are dominated by external references rather than intra-step contradictions: among critical errors with an agreed reference-category label, \textit{Task Conflict} and \textit{History Conflict} together cover $98.7\%$ of \tautwobench\ and $97.5\%$ of SWE-Bench Pro critical errors, while \textit{Intra-Step Conflict} accounts for only $1.3\%$ and $2.5\%$, respectively.
The two subsets differ in which external reference is violated more often: \tautwobench\ is balanced between \textit{Task Conflict} ($52.4\%$) and \textit{History Conflict} ($46.3\%$), reflecting failures that violate either explicit user constraints or earlier tool observations; SWE-Bench Pro is more strongly skewed toward \textit{Task Conflict} ($73.4\%$ vs. $24.1\%$ \textit{History Conflict}), consistent with code-task failures where the violated reference is most often the task instruction or expected behaviour rather than an intermediate observation.

\paragraph{Joint patterns.}
Reading the diagrams jointly, two patterns stand out.
First, the reasoning phase is the dominant entry channel into both \textit{Task Conflict} and \textit{History Conflict} in both subsets, suggesting that strengthening evidence-grounded reasoning is the single highest-leverage intervention for critical-error reduction.
Second, the planning phase flows almost exclusively into \textit{Task Conflict}, whereas observation-phase failures flow almost exclusively into \textit{History Conflict}; this clean separation reflects the underlying definitions and supports using the reference category as a complementary axis to the execution phase when analysing failure mechanisms.

\section{Case Studies}
\label{sec:appendix:case}

We present an end-to-end case to illustrate how \ourmas localizes the critical error within a long failed trajectory.
The example is drawn from the airline subset of \tautwobench, where \texttt{qwen3-max} acts as the agent and the user requests several modifications to an existing reservation under a budget cap of \$200.
The trajectory contains $49$ steps and ends with the agent transferring the conversation to a human operator.
\ourmas selects step $25$ as the critical error, matching the human annotation exactly.

\subsection{Case Overview}
\label{sec:appendix:case_overview}

\begin{table}[h]
\centering
\small
\caption{Overview of the case study trajectory.}
\label{tab:appendix:case_overview}
\begin{tabular}{ll}
\toprule
Field & Value \\
\midrule
Dataset             & \tautwobench (airline) \\
Agent model         & \texttt{qwen3-max} \\
Trajectory length   & 49 steps \\
Final outcome       & Failed (transferred to human) \\
Detected triggers   & 15 \\
Detected instances  & 9 \\
Candidate instances & 4 (in $\mathcal{F}(\tau)$) \\
\ourmas critical step      & 25 \\
Human-annotated step       & 25 \\
\bottomrule
\end{tabular}
\end{table}

\paragraph{Task.}
User \texttt{yara\_garcia\_1905} asks the agent to modify reservation \texttt{HXDUBJ}: change the outbound flight to a direct flight one day later, push the return flight back by one day, and, within a \$200 budget, consider upgrading to business class and adding checked baggage.
The system policy explicitly states that all flights in the same reservation must share the same cabin class.

\paragraph{Trajectory summary.}
The trajectory can be decomposed into three phases.
\textit{Flight modification (steps 0--18).}
The agent retrieves the reservation, searches for direct flights, and successfully updates the booking to the new outbound (\texttt{HAT072}) and return (\texttt{HAT278}) flights in economy.
\textit{Upgrade negotiation (steps 19--24).}
The agent computes the cost of upgrading both flights to business as \$408, which exceeds the user's \$200 budget; the user then asks whether only the outbound flight can be upgraded.
\textit{Cascading failure (steps 25--48).}
The agent commits to the partial-upgrade plan, executes the corresponding API call, and enters a chain of payment failures and repeated retries that finally end in a hand-off to a human agent.

\subsection{The Critical Error}
\label{sec:appendix:case_critical_step}

\begin{tcolorbox}[breakable, colback=red!3, colframe=red!50!black, fonttitle=\small\bfseries,
  title={Step 25 (assistant) --- Critical Error}]
\small
\textit{Upgrading only the outbound flight (HAT072) to business class costs \$282. Since your current outbound flight costs \$130, the upgrade requires an additional \$152. Given your budget of \$200, this upgrade is within your budget. As a gold member, you will still enjoy two free checked bags on both flights. Please confirm whether you would like to proceed with upgrading only the outbound flight to business.}
\end{tcolorbox}

The agent commits to a plan that violates the reservation-level uniform-cabin policy stated in the system prompt.
The trigger detector flags this step as a \textit{Task Conflict} in the \textit{planning} phase with the wrong commitment ``upgrade only the outbound flight to business'' and the violated reference being the policy ``all flights in the same reservation must share the same cabin class''.
Because this commitment cannot be repaired without retracting the plan, the resulting error instance has a \textit{semantic} terminal footprint and the state label \textit{Manifest Active}, which keeps it in the candidate set $\mathcal{F}(\tau)$.

\subsection{Detected Error Instances}
\label{sec:appendix:case_instances}

\ourmas clusters the $15$ per-step triggers into $9$ object-anchored error instances.
Table~\ref{tab:appendix:case_instances} lists them with origin step, state label, and a short description.
Four instances (origin steps $13$, $25$, $33$, $39$) survive state-based filtering and form the candidate set $\mathcal{F}(\tau)$.

\begin{table*}[t]
\centering
\small
\caption{Detected error instances in the case-study trajectory.
Instances marked with $\star$ form the candidate set $\mathcal{F}(\tau)$, and the row marked $\dagger$ is the critical instance selected by \ourmas.}
\label{tab:appendix:case_instances}
\begin{tabularx}{\textwidth}{cccccX}
\toprule
ID & Origin & Triggers & Reference category & State label & Description \\
\midrule
0 & 9  & 9              & Task Conflict       & Clean Resolution & Flight search ignores the user-specified 8\,am--9\,pm time window, but the next turn still yields a feasible option. \\
$1^{\star}$ & 13 & 13, 19, 23 & History Conflict & Costly Resolution (budget debt) & Financial computations omit the non-refundable \$30 insurance fee, propagating through the refund and upgrade-cost estimates. \\
2 & 15 & 15             & Task Conflict       & Clean Resolution & The agent incorrectly claims that travel insurance waives change fees. \\
3 & 21 & 21             & Task Conflict       & Clean Resolution & The agent queries user details when the budget is already obviously insufficient. \\
$4^{\star\dagger}$ & 25 & 25, 27, 31 & Task Conflict & Manifest Active (semantic) & Commits to upgrading only the outbound flight, violating the reservation-level uniform-cabin policy. \\
$5^{\star}$ & 33 & 33, 41         & Task Conflict       & Manifest Active (semantic) & Repeatedly proposes split payment in a single \texttt{update\_reservation} call, which the API does not support. \\
6 & 35 & 35             & History Conflict    & Clean Resolution & Attempts to pay with a certificate whose balance (\$150) is below the required amount. \\
7 & 37 & 37             & History Conflict    & Clean Resolution & Reuses a payment method that has already failed in a prior step. \\
$8^{\star}$ & 39 & 39, 43         & History Conflict    & Manifest Active (semantic) & Repeatedly retries the same failed \texttt{gift\_card\_1646646} payment without addressing the underlying cause. \\
\bottomrule
\end{tabularx}
\end{table*}

\subsection{Error Propagation and Attribution}
\label{sec:appendix:case_propagation}

Although several earlier instances contain genuine local mistakes, only the critical instance at step $25$ has a wrong commitment that directly causes the terminal failure.
The propagation path is:
(i) at step $27$, the agent issues \texttt{update\_reservation\_flights} with \texttt{cabin=business} but provides the return flight at its economy price, so the system charges both flights as business (\$282+\$443=\$725), exceeding the gift-card balance;
(ii) at step $31$, the agent switches to another gift card and reissues the same invalid call, which again fails;
(iii) the agent then misattributes the failure to a payment-side problem and enters a payment-retry chain (instances $5$ and $8$, steps $33$--$43$), exhausting the remaining budget before transferring to a human at step $45$.

Within the candidate set $\mathcal{F}(\tau)$, the causal attribution stage compares:
(a) instance $1$ (budget debt, origin step $13$): a financial-calculation error that wastes budget but does not by itself prevent task completion;
(b) instance $4$ (semantic, origin step $25$): a plan that violates a system constraint and makes the requested upgrade infeasible regardless of payment choice;
(c) instances $5$ and $8$ (semantic, origin steps $33$ and $39$): downstream payment-side errors that occur only because the agent is already trying to execute the impossible plan committed at step $25$.
Removing instance $4$ would eliminate instances $5$ and $8$ and let the agent either accept the over-budget upgrade or decline it cleanly, while removing instance $1$ would still leave the partial-upgrade commitment unresolved.
Step $25$ is therefore selected as the critical error step $t^{\star}$.

\subsection{Discussion}
\label{sec:appendix:case_discussion}

The case illustrates how the three stages of \ourmas support long-horizon failure analysis.
First, error trigger detection grounds local mistakes in explicit conflicts, such as the uniform-cabin policy violation at step $25$.
Second, error state classification groups repeated triggers into object-anchored instances and distinguishes corrected errors (e.g., instances $0$, $2$, $3$) from terminal-relevant states such as budget debt and manifest active commitments.
Third, causal attribution compares the remaining candidates by their role in the failure rather than by temporal order alone, which is why \ourmas selects the later step $25$ instead of the earlier financial error.

\section{\ourmas Implementation Details}
\label{sec:appendix:prompts}

All experiments use Qwen3-235B-A22B-Thinking~\citep{yang2025qwen3} as the base model, with temperature fixed at 0 for reproducibility.
Following the error-lifecycle tracking perspective in \S\ref{sec:method:lifecycle}, \ourmas operationalizes critical error detection through three stages: error trigger detection, error state classification, and causal attribution.
To handle long trajectories, we use the multi-granularity compression scheme in \S\ref{sec:method:compression}.
The high-detail view (th1) keeps up to 3000 characters, the medium-detail view (th2) keeps up to 1200 characters, and the low-detail view (th3) keeps up to 600 characters.

Each downstream stage uses these views according to its own evidence requirements.
In the \textit{error trigger detection} stage, the current step and history within two steps are rendered with th1, history three to five steps away is rendered with th2, and more distant history is rendered with th3.
Only the history before the current step is included in this view; the current step is provided separately in th1 for local verification.
In the \textit{error state classification} stage, trigger steps are rendered with th1 and all other steps are rendered with th3 as background context for instance clustering; for state classification, the origin step, the last trigger step, and a fixed number of following steps are always rendered with th1.
The remaining context is rendered by falling back in the order th1 $\to$ th2 $\to$ th3 $\to$ raw, with inline annotations indicating the associated error instance.
For \textit{budget debt}, we use a Python script to check whether the distance between the origin step and the repair step exceeds 50\% of the trajectory length.
In the \textit{causal attribution} stage, the full trajectory is rendered with the same fallback order th1 $\to$ th2 $\to$ th3 $\to$ raw, with an additional 800-character truncation limit.

We list the complete system prompts for the compression step and the three stages of \ourmas: error trigger detection, error state classification, and causal attribution.

\paragraph{Multi-Granularity Compression Prompt.}

\begin{tcolorbox}[breakable, colback=gray!5, colframe=black!40, fonttitle=\small\bfseries,
  title={Stage A --- Trajectory Compression}]
\begin{lstlisting}[basicstyle=\tiny\ttfamily, breaklines=true, columns=flexible,
  keepspaces=true, upquote=true]
You are compressing a single trajectory step at THREE different compression levels for downstream agent error diagnosis. Produce a detailed version (th1), a moderate version (th2), and a concise version (th3) of the same step content.

1. PRESERVE SEMANTIC CORE: Compression removes ONLY redundant phrasing, verbose boilerplate, repetitive expressions, and filler words. The core meaning of EVERY sentence must survive. When in doubt, keep more rather than less.

2. CONSTRAINTS AND PLANS ARE CRITICAL: If the content contains ANY of the following, their key information MUST be preserved across ALL three tiers:
   - Constraints (time limits, budget limits, "do NOT do X" prohibitions, "must do Y first" preconditions, required formats, allowed/disallowed actions)
   - Plans (specific action items, ordered steps, goals, sub-goals, milestones)
   - Requirements for subsequent steps (instructions that govern future behavior, expected outputs, acceptance criteria)
   - Conditional logic ("if X then Y", fallback strategies, error handling rules)
   Drop surrounding prose before dropping any constraint or plan item.

3. THREE-TIER COMPRESSION LEVELS:
   - th1 (detailed, <={th1_max} chars per field): Preserve all meaningful details. Remove only obvious redundancy, repeated phrasings, and decorative formatting. Keep every distinct fact, constraint, action, and observation.
   - th2 (moderate, <={th2_max} chars per field): Merge related information. Omit secondary details and elaborations. But KEEP all constraints, key decisions, action outcomes, error messages, and critical observations.
   - th3 (concise, <={th3_max} chars per field): Retain only the most essential actions, results, errors, and constraints. Use minimal wording. Still preserve all hard constraints and critical factual anchors.

================ PRESERVE VERBATIM =============

Preserve the FOLLOWING classes of tokens VERBATIM (copy them exactly as written - do not translate, paraphrase, reorder, round, or abbreviate):

- Proper nouns and named entities (people, places, brands, franchise/universe names such as "Disney", "Marvel", "MCU", book/paper/file names).
- Product / part / SKU / ASIN / ISBN / DOI / arXiv identifiers and any alphanumeric IDs the environment returned.
- Numbers with their units exactly as given (prices like "$22.50", quantities like "3 items", measurements like "45 min", percentages, counts).
- Years and dates (including citation years like "(1976)", ISO dates, "Q3 2023"-style period labels).
- URLs, file paths, CLI flags, API parameter names and their literal values.
- Quoted constraint phrases from the task statement (e.g. "avoid Tue", "refundable", "under $25", "first page only") - keep them inside quotation marks if the original had them.
- Tool / function names and their exact argument keys.
- Error messages, status codes, truncation markers ("has_more: true", "output truncated", "403 Forbidden").

You MAY rewrite the connective prose between these tokens to shorten the field; you MAY NOT rewrite the tokens themselves. If preserving them verbatim would push a field past the length cap, DROP the least-important surrounding prose first and keep the verbatim tokens.

\end{lstlisting}
\end{tcolorbox}

\paragraph{Error-Trigger Annotation Prompt.}

\begin{tcolorbox}[breakable, colback=gray!5, colframe=black!40, fonttitle=\small\bfseries,
  title={Stage 1 --- Error Trigger Detection}]
\begin{lstlisting}[basicstyle=\tiny\ttfamily, breaklines=true, columns=flexible,
  keepspaces=true, upquote=true]
You are an error-trigger annotator for one step of an LLM-agent trajectory. For the CURRENT STEP, decide which trigger tags (if any) fire. Each fired trigger falls into one of four categories:

  cat-1 = the step conflicts with the TASK
  cat-2 = the step conflicts with VISIBLE CONTEXT (env / tool output / system feedback / prior environment-rendered fact)
  cat-3 = the step is INTERNALLY INCONSISTENT - its claim contradicts something the same message states verbatim, or contradicts the agent's own prior plan / reflection / memory
  env   = the agent's tool call at THIS step is correct but the environment response is anomalous;

=== CHECKLIST (A) - conflict with TASK (cat-1) ===

  [ ] The step's plan permanently drops a sub-task / sub-goal / deliverable the TASK explicitly requires, with no TODO / "later" / sequencing intent in the step itself, and HISTORY does not already cover it. Phased execution that still keeps deferred items in scope is NOT a trigger.
      -> plan.BadDecomposition

  [ ] The step's plan schedules a sub-task before another sub-task whose output it depends on, and TASK itself dictates this ordering.
      -> plan.WrongOrder

  [ ] The step depends on a tool / API / permission / data source that TASK or the environment does not provide.
      -> plan.UnrealisticPlan

  [ ] Double-anchor required. (i) HISTORY contains a verbatim line where the agent was previously on-constraint; (ii) THIS step contains a verbatim line that explicitly narrows / widens / replaces a TASK constraint. Both anchors must be in reference_quote. Gradual multi-step drift without an explicit single-step swerve is NOT a trigger here.
      -> plan.GoalDrift

  [ ] TASK explicitly requires verify / check / confirm of X, and THIS step delivers the final answer without performing that check. Implicit verification expectations are NOT a trigger here.
      -> verify.NoVerification

  [ ] The step declares completion / returns the final answer / stops the trajectory while a TASK-required deliverable is still unmet.
      -> verify.PrematureTermination

  [ ] The step performs an action that TASK or environment rules explicitly forbid (safety policy, permission boundary, forbidden endpoint, forbidden operation).
      -> act.UnsafeOrForbiddenAction

=== CHECKLIST (B) - conflict with VISIBLE CONTEXT (cat-2) ===

  [ ] The step makes an explicit decision and cites supporting evidence, but the citation is distorted (wrong value, wrong field, mis-paraphrased), and the wrong reading drives the decision.
      -> reason.WrongChoice

  [ ] The step cites visible evidence correctly, but generalises the conclusion beyond what the evidence supports.
      -> reason.InvalidInference

  [ ] The step relies on a prior tool call / observation / fact that does NOT actually appear in HISTORY (fabricated cross-message reference into a context that should exist).
      -> reason.MissingAssumptionCheck

  [ ] (a) The chosen tool or data source is unsuited to the established sub-goal; OR (b) plan-action mismatch: this step commits to plan A but issues action B.
      -> act.WrongTool

  [ ] A tool call is syntactically malformed against the visible schema or prior successful calls.
      -> act.ToolSchemaMismatch

  [ ] The action order violates a workflow established by HISTORY or the visible UI / env state machine.
      -> act.WrongActionSequence

  [ ] The step misreads its own immediate tool / env output: treats an error as success, swaps fields, mis-identifies the returned entity.
      -> obs.MisreadOutput

  [ ] Subject to (C4): information visible in current or prior tool / env output is silently ignored where it is clearly needed, OR a summary/memory block drops load-bearing facts.
      -> obs.IgnoreOutput

  [ ] The step treats a partial / pending / truncated response as final.
      -> obs.TimingIssue

  [ ] The step binds intent to the wrong visible entity.
      -> obs.GroundingFail

  [ ] The step is performing verification but the criterion does not match the sub-goal in HISTORY.
      -> verify.WrongVerification

  [ ] The step retries the same tool / query / action that already produced bad / empty / error in HISTORY, with no relevant env state change and no strategy change. Subject to (C4).
      -> verify.InfiniteRetry

=== CHECKLIST (C) - internally inconsistent (cat-3) ===

  [ ] The conclusion contradicts something the SAME message states verbatim, OR contradicts a prior agent reflection / memory / claim.
      -> reason.InvalidInference

  [ ] The step picks the wrong row / column / item from a list / table that is verbatim visible in the CURRENT MESSAGE itself.
      -> obs.MisreadOutput

  [ ] The step asserts a concrete factual claim from the agent's parametric knowledge, where the claim is concrete/falsifiable, NOT supported by TASK/HISTORY/CURRENT MESSAGE, clearly wrong on widely-known ground truth, and load-bearing for this step's conclusion.
      -> reason.MissingAssumptionCheck

=== CHECKLIST (D) - environment-side anomaly (env) ===

env triggers fire when the agent's tool call at this step is otherwise correct, but the immediately following environment response shows the anomaly. Set attribution = "env".

  [ ] The search / page response is hijacked by an ad vignette / overlay.
      -> env.AdOverlayHijack

  [ ] The model API / platform content filter rejects a syntactically well-formed query.
      -> env.ContentFilterBlock

  [ ] The tool call is admissible but the returned observation is structurally insufficient because the tool is degenerate or blind on this target.
      -> env.ToolExtractorDegenerate

  [ ] The observation contains verbatim evidence of HTTP 429 / 5xx / timeout / connection reset / rate limited.
      -> env.RateLimitOrTransient

  [ ] The tool returns structurally empty / constant / payload identical to a prior call.
      -> env.EmptyOrRepeatedPayload

=== CONFIDENCE ===

For each trigger fill `confidence`:
  - high   : (C1)+(C2) are both crisp; wrong_content_quote and reference_quote are unambiguous and the conflict is direct.
  - medium : both quotes exist verbatim, but reference_quote needs light alignment.
  - low    : inferring from a pattern; verbatim anchors are weak or partial.
Always fill `confidence_reasoning` with 1-2 sentences.

 Two verbatim quotes per trigger:
 - wrong_content_quote : verbatim substring of the CURRENT STEP carrying the wrong content;
 - reference_quote     : verbatim substring of TASK / HISTORY / CURRENT STEP that is the I being conflicted.
 Paraphrase, summary, or composite quotes are not acceptable. If either quote cannot be cited verbatim, do NOT emit the trigger.
     
=== OUTPUT FORMAT ===

Emit a single JSON object:

{
  "triggers": [
    {
      "step": <int>,
      "category": "cat-1" | "cat-2" | "cat-3" | "env",
      "taxonomy_tag": "<one of the 25 tags listed in the checklists above>",
      "attribution": "agent" | "env",
      "wrong_content_quote": "<verbatim from CURRENT STEP, or for the upstream-env back-reference verbatim from the earlier HISTORY step>",
      "reference_quote": "<verbatim from TASK / HISTORY / CURRENT STEP>",
      "confidence": "high" | "medium" | "low",
      "confidence_reasoning": "<1-2 sentences>"
    }
  ]
}
\end{lstlisting}
\end{tcolorbox}

\paragraph{Error-Instance Clustering Prompt.}

\begin{tcolorbox}[breakable, colback=gray!5, colframe=black!40, fonttitle=\small\bfseries,
  title={Stage 2 --- Error State Classification: Instance Clustering}]
\begin{lstlisting}[basicstyle=\tiny\ttfamily, breaklines=true, columns=flexible,
  keepspaces=true, upquote=true]
You are an error-instance clustering annotator. You receive a list of per-step error TRIGGERS detected on one trajectory. Group those triggers into ERROR INSTANCES, where one instance corresponds to "the same underlying erroneous content / behavior" repeated or reflected across one or more steps.

=== TERMINOLOGY ===

Each trigger contradicts a "conflict object" - the thing the wrong content is conflicting with. Following the spec we call this object I. Concretely, I can be one of:
  - a TASK clause (sub-task, deliverable, constraint);
  - a HISTORY object (a specific physical / virtual location, a specific tool output, a specific observation, a specific memory entry, a specific prior tool-call signature);
  - a SAME-MESSAGE premise (operands, listed items, restated premise);
  - a prior agent-self statement (plan, reflection, memory).

Two triggers are about the SAME I when they are conflicting with the same underlying thing.
Two triggers are about DIFFERENT I's when they conflict with different observations / different TASK clauses / different self-claims, or happen to share the taxonomy_tag string but the underlying object is different.

=== CLUSTERING RULES ===

(I-a) Same step, same I, multiple modules -> ONE instance.
(I-b) Same step, DIFFERENT I -> MULTIPLE instances.
(I-c) Across steps, same I, never repaired in between -> ONE instance with origin_step = first occurrence.
(I-d) Across steps, same I, but repaired and then re-fires later -> SPLIT into TWO instances. If you cannot determine whether a repair happened, conservatively keep them in ONE instance.
(I-e) "Fabricated I" (an I the agent invented out of thin air) is identified by the verbatim claim that introduced it.

CLUSTERING PRINCIPLE: cluster by the SPECIFIC CONSTRAINT OBJECT I being violated, NOT by "same type of erroneous behavior".

=== PRECISION REQUIREMENTS FOR `what_is_being_violated` ===

Write `what_is_being_violated` as a CONCRETE pointer to the exact object being contradicted.
  - For TASK-clause violations: quote or paraphrase the specific constraint / deliverable / format requirement.
  - For HISTORY-object violations: quote or paraphrase the specific observation / tool output / location / URL, and reference the step where it first appeared.
  - For SAME-MESSAGE premise violations: name the operands / listed items being misused.
  - For prior-self violations: quote or paraphrase the prior plan / reflection / memory statement being contradicted.

=== ONE-SHOT PROBE GUIDANCE ===

If a trigger fires at exactly ONE step and the agent at later steps neither references the same wrong belief nor takes another action against the same I, keep it as a STANDALONE singleton instance. Do NOT merge it with other unrelated triggers.

=== HARD CONSTRAINT ===

Triggers with attribution="agent" and triggers with attribution="env" CANNOT be merged into the same instance.
Every input trigger index must end up in EXACTLY ONE instance's trigger_indices.

=== ENV INSTANCE MERGING ===

For triggers with attribution="env", apply same-source merging when two env triggers share at least one of:
  - Same registrable domain;
  - Same tracking parameter key+value;
  - Same HTTP error code;
  - Same failure keyword: "rate limit", "timeout", "content filter", "empty payload", "captcha", "access denied", "not found";
  - Same taxonomy_tag.

Do NOT merge env triggers that share only a generic/CDN domain (google.com, bing.com, youtube.com, etc.).

=== TRAJECTORY CONTEXT ===

You will also receive a TRAJECTORY STEP CONTENT block. Steps that fired at least one trigger are shown at HIGH detail (th1 compression); steps that did not fire any trigger are shown at LOW detail (th3 compression) as background for later state classification.

=== OUTPUT FORMAT ===

{
  "instances": [
    {
      "instance_id": <int, 0-indexed across this trajectory>,
      "trigger_indices": [<int>, ...],
      "origin_step": <int, smallest step among trigger_indices>,
      "last_trigger_step": <int, largest step among trigger_indices>,
      "attribution": "agent" | "env",
      "category": "cat-1" | "cat-2" | "cat-3" | "env",
      "error_content": "<one-sentence natural-language description of the same erroneous content/behavior>",
      "what_is_being_violated": "<one-sentence description of the I being violated>",
      "merged_reasoning": "<2-3 sentences explaining why these triggers share the same I>"
    }
  ]
}

If the input triggers array is empty, emit  {"instances": []}.
\end{lstlisting}
\end{tcolorbox}

\paragraph{Error State Classification Prompt.}

\begin{tcolorbox}[breakable, colback=gray!5, colframe=black!40, fonttitle=\small\bfseries,
  title={Stage 2 --- Error State Classification: State Labeling}]
\begin{lstlisting}[basicstyle=\tiny\ttfamily, breaklines=true, columns=flexible,
  keepspaces=true, upquote=true]
You are an instance-state annotator. You receive ONE error instance and the FULL trajectory. Upstream trigger detection is high-recall and may flag exploratory or weakly committal steps; your job is to perform the commitment-strength recheck, the repair / state determination, and the terminal-connection audit defined in the methodology document.

=== KEY DEFINITIONS ===

  origin_step       = the first trigger step selected for this instance. 
  qualified_origin_step = the FIRST step in [origin_step, T] at which the agent makes an OBSERVABLE WRONG COMMITMENT for this instance's violated_object. May equal origin_step, may be a later step, may be null.
  last_trigger_step = the latest step at which the same instance re-fired upstream.
  terminal step T   = the last message index in the trajectory.
  I (violated_object) = the concrete object the instance is violating (a TASK clause / a HISTORY object / a same-message premise / a prior agent-self statement).

=== TASK 1 - COMMITMENT-STRENGTH RECHECK ===

Upstream trigger detection is high-recall: a trigger may fire at an early probe or initial plan step even when the agent had no way yet to know the action was wrong. Your job is to classify each origin_step into one of three commitment strengths so that `qualified_origin_step` reflects the FIRST step at which the agent observably persisted in the wrong belief despite already-visible evidence.

A commitment to error requires BOTH:
  (a) the agent takes an observable binding action or states a non-hedged belief that contradicts the violated_object I, AND
  (b) by that step, the evidence contradicting I is already visible in the prior trajectory.

Classify the upstream origin into ONE of:

  origin_commitment_status = "explicit_wrong_commitment"
    BOTH (a) and (b) hold at origin_step. Set qualified_origin_step = origin_step.

  origin_commitment_status = "weak_signal"
    EITHER (a) holds at origin_step but (b) does NOT (conflicting evidence not yet observable), OR (a) is only hedged at origin_step.
    There MUST be a clearly LATER step M (origin_step < M <= T) at which the agent, now able to see the conflicting evidence, still takes a binding action that contradicts I.
    Set qualified_origin_step = M, set origin_relocated = true.

  origin_commitment_status = "pure_exploration"
    origin_step is plain exploration / setup / a reasonable first guess, AND no step in (origin_step, T] satisfies the weak_signal relocation threshold.
    Set qualified_origin_step = null, set exploration_suppressed = true.

SINGLE-TRIGGER SHORTCUT. If the instance has exactly ONE trigger (origin_step == last_trigger_step) AND the instance is cat-2 or cat-3 (NOT cat-1 or env), the default label is `pure_exploration`, UNLESS a verbatim quote from a step strictly after origin_step shows the agent referencing or committing to the same wrong object.

  HARD CAT-1 RULES:
    1. cat-1 instance MUST NOT have origin_commitment_status = pure_exploration.
    2. cat-1 instance MUST NOT have state = dormant.

  HARD ENV RULES:
    1. An env instance MUST NOT have origin_commitment_status = pure_exploration UNLESS BOTH: (a) the environment self-recovers on the next judgable step, AND (b) the agent's next action is consistent with the recovered output.
    2. For env instances, wasted steps should be counted against the VIOLATED sub-goal object remaining unresolved.

=== TASK 2 - REPAIR / STATE DETERMINATION ===

  fixed_at_step_<N>
    There is a step N with anchor < N <= T at which the agent's behavior satisfies the repair criterion for this instance's category.

  cat-1 : at step N the agent touches the violated TASK clause AND the behavior is consistent with that constraint.
  cat-2 : at step N the agent touches the violated CONTEXT object AND the behavior is consistent with the latest visible state.
  cat-3 : at step N the agent explicitly retracts or corrects the prior wrong claim.
  env   : (env-fix-1) the agent detects the env anomaly and switches to a MATERIALLY DIFFERENT strategy; OR (env-fix-2) the env recovers AND the agent's subsequent behavior is consistent with the recovered output.

  HARD ENV-FIX RULES:
    (R1) The following are NOT a strategy switch: clicking back; retrying the same URL/query/tool; reloading the same hijacked/errored page.
    (R2) Same-source RE-FIRE invalidates a prior fix.
    (R3) A genuine env-fix-1 requires BOTH (a) the agent explicitly recognises the env anomaly, AND (b) the next observable action targets a DIFFERENT path.

=== TASK 3 - TERMINAL-CONNECTION AUDIT (FAILURE-PATH TEST) ===

Decide whether THIS instance lies on the actual observable path that took the trajectory to its terminal failure.

  FAILURE-PATH MEMBERSHIP. Does any of the following appear on the path that produced the observed terminal failure?
         (i)   the instance's wrong commitment is still active at T, OR re-fires after qualified_origin_step;
         (ii)  the instance is reflected in the terminal answer, terminal action, terminal failure state;
         (iii)  this instance produced an irreversible state change visible at T.
       If ANY of (i)-(iii) holds -> the instance IS on the failure path.

Decide terminal_connection, one of:
  "irreversible"    - Q1(iv) holds.
  "semantic"        - Q1(i) or Q1(ii) holds.
  "none"            - Q2 fully passes, or Q1 has no evidence at all.

=== TASK 4 - RESOURCE EFFECT ===

Fill `resource_effect` whenever the instance wasted ANY steps in (qualified_origin_step, T].

Resource effect schema:
  {
    "wasted_steps": [<list of step indices wasted on I>],
    "wasted_step_count": <int>,
    "last_referencing_step": <int or null>
  }

=== OUTPUT FORMAT ===

Return STRICT JSON with EXACTLY these keys:

{
  "instance_id": <int, copy from input>,
  "origin_commitment_status": "explicit_wrong_commitment" | "weak_signal" | "pure_exploration",
  "qualified_origin_step": <int or null>,
  "fix_status": "active" | "fixed_at_step_<N>",
  "fix_evidence_quote": "<verbatim quote or null>",
  "terminal_connection": "irreversible" | "semantic" | "budget_debt" | "none",
  "chain_membership": <bool>,
  "resource_effect": <object or null>,
  "chain_explanation": "<2-4 sentences. MUST include: (1) the trajectory's terminal failure mode in one phrase, (2) which of Q1(i)-(iii) holds OR the named dominating cause (3) one verbatim evidence quote.>"
}
\end{lstlisting}
\end{tcolorbox}

\paragraph{Causal Attribution Prompt.}

\begin{tcolorbox}[breakable, colback=gray!5, colframe=black!40, fonttitle=\small\bfseries,
  title={Stage 3 --- Causal Attribution}]
\begin{lstlisting}[basicstyle=\tiny\ttfamily, breaklines=true, columns=flexible,
  keepspaces=true, upquote=true]
You are an expert at identifying the SINGLE critical step that caused a failed agent trajectory.

An error is critical if:
- It represents the ROOT CAUSE that made task success impossible
- It caused a cascade of subsequent errors
- The trajectory could have succeeded if THIS specific error had not occurred
- IMPORTANT: Correcting this specific error would fundamentally change the trajectory toward success

# Hints
1. Consider the ENTIRE trajectory from a global perspective - understand the task goal and how the agent's path diverged from success.
2. Early exploration steps (steps 1-3) are often normal and should NOT be marked as critical unless they clearly introduce the root cause.
3. Prefer the first step where the agent had enough reasonable information to proceed correctly but nevertheless introduced the error locally.
   This includes producing an incorrect output, misinterpreting or misusing prior information, or turning a correct intermediate state into an incorrect one.
4. Do not choose a later step merely because the failure becomes more visible there.
   If a later step only repeats, propagates, or amplifies an earlier mistake, select the earlier step where the mistake originated.
5. If a flawed strategy is repeated across multiple steps, attribute the failure to the step where that strategy was first introduced. If the strategy was reasonable but the execution result was wrong, attribute the failure to the first execution step that produced the incorrect result.
6. Use recoverability only as a tie-breaker: among otherwise similar candidates, prefer the earliest step whose error made recovery unlikely or blocked the trajectory from returning to a successful path.

You are given:
- the TASK,
- the FULL TRAJECTORY,
- a list of SUSPICIOUS STEPS that an upstream annotator flagged as containing some erroneous content (they are hints; you may pick any other step from the trajectory if you disagree).

=== WHAT TO OUTPUT ===
Output ONE JSON object with these keys and nothing else:
- "critical_step": <int>  - the step number of the critical error.
- "rationale": <string, 1-3 sentences> explaining why this step is critical.

=== REQUIRED OUTPUT FORMAT (single JSON object, no prose around it) ===
{
  "critical_step": <int>,
  "rationale": "<1-3 sentences>"
}

Return ONLY the JSON object.
\end{lstlisting}
\end{tcolorbox}

\section{Experiment Details}
\label{sec:appendix:experiment}

All experiments were conducted with temperature set to 0 to ensure reproducibility.

\subsection{Direct Prompting Baselines}

To satisfy context-length constraints, we truncate each trajectory step to 2{,}048 characters and set the maximum context length to 16{,}384 tokens for each trajectory.
Table~\ref{tab:appendix:context_budget} evaluates whether this per-step limit disadvantages direct prompting.
Increasing the limit substantially lowers the proportion of truncated steps but does not improve accuracy, suggesting that retaining more raw detail can also introduce irrelevant content that interferes with critical-step judgment.

\begin{table}[h]
\centering
\small
\caption{Sensitivity of direct prompting to the per-step character limit.}
\label{tab:appendix:context_budget}
\begin{tabular}{lcc}
\toprule
Per-step limit & Avg.\ accuracy & Truncated steps \\
\midrule
2{,}048 characters & $\mathbf{26.00}$ & $27.1\%$ \\
4{,}096 characters & $20.02$ & $14.5\%$ \\
8{,}192 characters & $24.73$ & $3.5\%$ \\
\bottomrule
\end{tabular}
\end{table}

\begin{tcolorbox}[breakable, colback=gray!5, colframe=black!40, fonttitle=\small\bfseries,
  title={Direct Prompting Baseline --- Critical Error Detection}]
\begin{lstlisting}[basicstyle=\tiny\ttfamily, breaklines=true, columns=flexible,
  keepspaces=true, upquote=true]
You are an expert at analyzing failed agent trajectories and providing actionable debugging insights.

=== CRITICAL ERROR DEFINITION ===
A "critical error step" is the EARLIEST error instance that remains ACTIVE (unrepaired) at the terminal step of a failed trajectory -- the first error whose effects were never truly repaired, making it the root cause of the failure chain.

An error is "critical" if:
- It represents the ROOT CAUSE that made task success impossible
- It caused a cascade of subsequent errors
- The trajectory could have succeeded if THIS specific error had not occurred
- Correcting this specific error would fundamentally change the trajectory toward success

=== TASK ===
{task_description}

TRAJECTORY OUTCOME: FAILED (reward=0, reported by the task environment).
LAST STEP INDEX: {last_step}

=== FULL TRAJECTORY (one entry per message) ===
{trajectory_block}

Your task: find the critical error step of the trajectory.

=== REQUIRED OUTPUT FORMAT (single JSON object, no prose around it) ===
{
  "critical_error_analysis": {
    "step": <int>,
    "reason": "<1-3 sentences explaining why this step is a strong root-cause candidate>"
  }
}

Return ONLY the JSON object.
\end{lstlisting}
\label{tab:appendix:direct_prompting_baseline}
\end{tcolorbox}

\subsection{Multi-Agent Systems Baselines}
We reproduced each multi-agent baseline using its official codebase and default configuration.
To ensure a fair comparison with our implementation of \ourmas, we used Qwen3-235B-A22B-Thinking as the base model and fixed the temperature at 0 for reproducibility.

\subsection{Additional Evaluation Results}

\paragraph{Relaxed critical-step localization.}
Exact-step accuracy is deliberately strict, although predictions near the annotated critical step can still be useful in practice.
For automated intervention, a prediction after the critical error may arrive too late; we therefore additionally evaluate whether the prediction falls in the asymmetric window $[\mathrm{GT}{-}3,\mathrm{GT}]$.
As shown in Table~\ref{tab:appendix:relaxed_results}, \ourmas\ reaches the best average accuracy of $50.2\%$ and remains strongest on ALFWorld, \tautwobench, and SWE-Bench Pro.
For human-in-the-loop debugging, where nearby later steps can also guide inspection, \ourmas\ reaches $72.8\%$ under the symmetric $[\mathrm{GT}{-}5,\mathrm{GT}{+}5]$ window.

\begin{table*}[t]
\centering
\small
\adjustbox{max width=\textwidth}{%
\begin{tabular}{lcccccccc}
\toprule
\textbf{Method} & ALFWorld & GAIA & WebShop & W\&W-HC & W\&W-Alg. & \tautwobench & SWE-Bench Pro & AVG \\
\midrule
GLM-5.1           & $39.0$ & $42.9$ & $42.0$ & $22.4$ & $73.8$ & $51.2$ & $12.8$ & $40.6$ \\
Gemini-3.1-Pro    & $38.0$ & $34.7$ & $40.0$ & $29.3$ & $62.7$ & $65.5$ & $23.3$ & $41.9$ \\
DeepSeek-V4-Pro   & $35.0$ & $63.3$ & $40.0$ & $29.3$ & $66.7$ & $56.8$ & $17.4$ & $44.1$ \\
Claude Sonnet 4.6 & $44.0$ & $42.9$ & $\mathbf{48.0}$ & $20.7$ & $54.0$ & $53.0$ & $23.3$ & $40.8$ \\
GPT-5.4           & $38.0$ & $61.2$ & $38.0$ & $22.4$ & $54.8$ & $51.5$ & $15.1$ & $40.1$ \\
Claude Opus 4.6   & $36.0$ & $44.9$ & $\mathbf{48.0}$ & $32.8$ & $69.0$ & $53.5$ & $19.8$ & $43.4$ \\
Qwen3-235B-A22B   & $37.0$ & $46.9$ & $46.0$ & $29.3$ & $69.0$ & $56.5$ & $25.6$ & $44.3$ \\
\midrule
AgentDebugger      & $45.0$ & $\mathbf{64.0}$ & $44.4$ & $29.3$ & $\mathbf{72.2}$ & $52.5$ & $16.3$ & $46.2$ \\
CHIEF              & $12.0$ & $48.0$ & $22.0$ & $\mathbf{34.5}$ & $65.1$ & $42.6$ & $19.8$ & $34.8$ \\
AgentRX            & $19.4$ & $32.7$ & $26.5$ & $31.0$ & $59.2$ & $52.2$ & $8.3$  & $40.8$ \\
\midrule
\textbf{\ourmas\ (Ours)} & $\mathbf{50.0}$ & $59.2$ & $40.0$ & $29.3$ & $69.0$ & $\mathbf{69.0}$ & $\mathbf{34.9}$ & $\mathbf{50.2}$ \\
\bottomrule
\end{tabular}%
}
\caption{Critical-step localization accuracy (\%) under the automated-intervention window $[\mathrm{GT}{-}3,\mathrm{GT}]$. AVG is the macro-average over the seven subsets.}
\label{tab:appendix:relaxed_results}
\end{table*}

\paragraph{Inference cost and compute-matched comparison.}
Table~\ref{tab:appendix:token_cost} reports average token consumption per trajectory using Qwen3-235B-A22B-Thinking as the common backbone.
To control for inference budget, we also run direct prompting $40$ times at temperature $1$ and use majority voting.
This baseline consumes slightly more tokens than \ourmas\ but reaches only $29.56\%$ accuracy, showing that \ourmas's gain over direct prompting cannot be explained by additional token consumption alone.

\begin{table}[h]
\centering
\small
\adjustbox{max width=\linewidth}{%
\begin{tabular}{lcc}
\toprule
\textbf{Method} & \textbf{AVG Accuracy} & \textbf{Avg.\ Tokens} \\
\midrule
Direct Prompting       & $25.69$ & $34{,}530$ ($1.0\times$) \\
\quad + Majority Voting & $29.56$ & $1{,}403{,}850$ ($40.6\times$) \\
AgentDebugger          & $23.72$ & $1{,}398{,}333$ ($40.5\times$) \\
CHIEF                  & $18.77$ & $307{,}786$ ($8.9\times$) \\
AgentRX                & $23.10$ & $218{,}862$ ($6.3\times$) \\
\textbf{\ourmas\ (Ours)} & $\mathbf{34.11}$ & $1{,}382{,}243$ ($40.0\times$) \\
\bottomrule
\end{tabular}%
}
\caption{Accuracy (\%) and token consumption. Majority Voting is a compute-matched direct-prompting baseline.}
\label{tab:appendix:token_cost}
\end{table}

\subsection{Application Experiment Details}
\label{sec:appendix:application}

\begin{tcolorbox}[breakable, colback=gray!5, colframe=black!40, fonttitle=\small\bfseries,
  title={Application --- Vanilla Prompt}]
\begin{lstlisting}[basicstyle=\tiny\ttfamily, breaklines=true, columns=flexible,
  keepspaces=true, upquote=true]
Trajectory:
{trajectory}

Your task:
1. Identify the earliest step which directly leads the agent off
   track or repeats ineffective behaviour.
2. Reference that exact step number as shown in the trajectory.
   Do not shift to later steps of that error.
3. Explain why the chosen step is wrong, citing relevant
   observation/action details.
4. Suggest a concrete alternative for that same step that would
   move the agent toward success
   (e.g., a specific action to take instead).

Respond strictly in the following format (single spaces around
colons, no extra text):
step:<number>
reason:<one concise, specific sentence>
suggestion:<one actionable suggestion for that step>
\end{lstlisting}
\end{tcolorbox}

\begin{tcolorbox}[breakable, colback=gray!5, colframe=black!40, fonttitle=\small\bfseries,
  title={Application --- Self-Reflection Prompt}]
\begin{lstlisting}[basicstyle=\tiny\ttfamily, breaklines=true, columns=flexible,
  keepspaces=true, upquote=true]
Current result: {trajectory}

Why is this trajectory not finished the task?

Feedback:
\end{lstlisting}
\end{tcolorbox}

\paragraph{Format-matched feedback comparison.}
The main application experiment compares complete debugging pipelines, whose feedback formats may differ.
To isolate the effect of critical-step localization, we additionally use the same repair-hint generator and the same injection procedure for Vanilla Debug and \ourmas; the two conditions differ only in the critical step supplied to the hint generator.
\ourmas's intermediate trigger, state, and attribution signals are hidden from both the hint generator and the actor.
Table~\ref{tab:appendix:format_matched_application} shows that \ourmas\ remains stronger across all three settings, providing a controlled comparison of the downstream value of its localized critical step.

\begin{table}[h]
\centering
\small
\adjustbox{max width=\linewidth}{%
\begin{tabular}{lccc}
\toprule
\textbf{Method} & \textbf{Airline} & \textbf{Retail} & \textbf{SWE-Bench} \\
\midrule
Initial & $78.00$ & $84.21$ & $72.00$ \\
Vanilla Debug & $86.00$ {\footnotesize($+8.00$)} & $90.35$ {\footnotesize($+6.14$)} & $78.00$ {\footnotesize($+6.00$)} \\
\textbf{\ourmas} & $\mathbf{88.00}$ {\footnotesize($\mathbf{+10.00}$)} & $\mathbf{92.98}$ {\footnotesize($\mathbf{+8.77}$)} & $\mathbf{80.00}$ {\footnotesize($\mathbf{+8.00}$)} \\
\bottomrule
\end{tabular}%
}
\caption{Task success rate (\%) in the format-matched per-trajectory repair experiment.}
\label{tab:appendix:format_matched_application}
\end{table}

\subsection{\ourmas Error Analysis}
\label{sec:error_analysis}

\paragraph{Candidate retention and conditional attribution.}
We first stratify all $869$ evaluated trajectories according to whether the human-annotated critical error is retained in the candidate set passed to final attribution; one of the $50$ GAIA trajectories is excluded because its ground-truth critical-error label is null.
As shown in Table~\ref{tab:appendix:candidate_analysis}, \ourmas\ retains the ground-truth step in $42.0\%$ of trajectories, compared with an expected $14.2\%$ coverage from a size-matched random candidate set.
When the ground truth is retained, candidate-guided attribution reaches $45.5\%$ micro-average accuracy, substantially outperforming direct holistic prompting at $29.6\%$.
When it is excluded upstream, \ourmas's evidence-constrained out-of-set fallback outperforms direct holistic prompting ($39.1\%$ versus $35.1\%$).
The fallback must identify an evidence-backed missing trigger together with its violated reference, origin step, and terminal footprint.
The overall $41.8\%$ reported here is a trajectory-level micro-average; Table~\ref{tab:main_results} reports a macro-average across the seven benchmark subsets.

\begin{table*}[t]
\centering
\small
\adjustbox{max width=\textwidth}{%
\begin{tabular}{lccccc}
\toprule
\textbf{GT status before attribution} &
\textbf{Cases (\%)} &
\textbf{Random coverage} &
\textbf{\ourmas} &
\textbf{Direct holistic} &
\textbf{$\Delta$} \\
\midrule
Retained in candidate set & $365$ ($42.0\%$) & $14.2\%$ & $166$ ($45.5\%$) & $108$ ($29.6\%$) & $+15.9$ pp \\
Excluded upstream         & $504$ ($58.0\%$) & $85.8\%$ & $197$ ($39.1\%$) & $177$ ($35.1\%$) & $+4.0$ pp \\
\midrule
Overall                   & $869$             & --       & $363$ ($41.8\%$) & $285$ ($32.8\%$) & $+9.0$ pp \\
\bottomrule
\end{tabular}%
}
\caption{Candidate-set coverage and conditional exact-step accuracy. Accuracy is micro-averaged within each stratum. Random coverage is the expected coverage of a uniformly sampled candidate set with the same size as \ourmas's candidate set.}
\label{tab:appendix:candidate_analysis}
\end{table*}

\paragraph{Stage-wise failure decomposition.}
\begin{table}[t]
  \centering
  \adjustbox{max width=\linewidth}{%
  \begin{tabular}{lccc}
  \toprule
  \textbf{Dataset} & \textbf{Trigger (\%)} & \textbf{State (\%)} & \textbf{Attribution (\%)} \\
  \midrule
  ALFWorld & $32.4$ & $6.8$ & $60.8$ \\
  GAIA & $20.0$ & $6.7$ & $73.3$ \\
  WebShop & $18.4$ & $0.0$ & $81.6$ \\
  WhoAndWhen-HC & $33.3$ & $2.2$ & $64.4$ \\
  WhoAndWhen-Algo & $62.1$ & $12.1$ & $25.9$ \\
  SWE-Bench Pro & $45.2$ & $24.2$ & $30.6$ \\
  \tautwobench & $49.2$ & $30.7$ & $20.1$ \\
  \midrule
  \textbf{Overall} & $\mathbf{42.1}$ & $\mathbf{17.7}$ & $\mathbf{40.1}$ \\
  \bottomrule
  \end{tabular}%
  }
\caption{Failure decomposition of \ourmas. }
\label{tab:error_analysis}
\end{table}

We decompose the failure cases of \ourmas\ according to the stage at which the ground-truth critical error is lost:
\textbf{Trigger Miss}, where the critical error is not identified as an error trigger;
\textbf{State Miss}, where the critical error is detected but filtered out from $\mathcal{F}(\tau)$ during error state classification;
and \textbf{Attribution Miss}, where the correct candidate enters the final attribution stage but is not selected.
Table~\ref{tab:error_analysis} shows that Trigger Misses ($42.1\%$) and Attribution Misses ($40.1\%$) are the two dominant failure modes, while State Misses are less frequent ($17.7\%$).

This decomposition characterizes where the remaining failures occur.
First, Trigger Misses ($42.1\%$) account for the largest share of failures. Although our trigger detector inspects each step under multiple compressed trajectory views and identifies evidence-grounded conflicts, some errors or deviations may be subtle, which can lead the detector to miss the critical error.
Second, the relatively low share of State Misses ($17.7\%$) suggests that error state classification usually preserves the critical error once it has been detected as a trigger. Attribution Misses ($40.1\%$) therefore correspond to harder residual cases after candidate filtering, where multiple retained errors are plausibly related to the final failure and the model must distinguish the earliest failure-responsible step.
These patterns suggest that further gains may come from improving recall for subtle trigger errors and refining attribution among already filtered, terminal-relevant candidates.